\documentclass[manuscript,screen]{acmart}

\pdfoutput=1

\usepackage{xr}
\usepackage{amsmath}
\usepackage{pifont}

\usepackage{subfig}
\usepackage{comment}
\usepackage{soul}
\usepackage{algorithmic}
\usepackage{hyperref}
\usepackage{booktabs}
\usepackage{threeparttable}
\usepackage{mathrsfs}
\usepackage{amsfonts}
\usepackage[ruled]{algorithm2e}

\usepackage{multirow}
\usepackage{graphicx}
\usepackage{color}
\usepackage{balance}
\usepackage{caption}
\usepackage{adjustbox}
\usepackage{soul}
\setstcolor{red}
\usepackage{array}
\usepackage{utfsym}
\usepackage{multirow}
\usepackage{enumitem}
\usepackage{colortbl}
\usepackage{xcolor}
\usepackage{wrapfig}

\usepackage{color}
\usepackage{tikz}
\usepackage[edges]{forest}
\definecolor{hidden-draw}{RGB}{0,0,0}
\definecolor{hidden-pink}{rgb}{0.98, 0.94, 0.75}
\definecolor{level0}{rgb}{0.67, 0.88, 0.69}
\definecolor{level1}{rgb}{0.98, 0.92, 0.84}
\definecolor{level2}{rgb}{0.8, 0.8, 1.0}
\definecolor{level3}{rgb}{1.0, 0.71, 0.76}
\definecolor{level4}{rgb}{0.49, 0.99, 0.0}

\newcommand{\definitionautorefname}{Definition}

\newcommand*{\shortautoref}[1]{%
  \begingroup
    \def\sectionautorefname{Sec.}%
    \def\subsectionautorefname{Sec.}%
    \def\figureautorefname{Fig.}%
    \def\tableautorefname{Tab.}%
    \def\equationautorefname{Eq.}%
    \def\subfigureautorefname{Fig.}%
    \def\definitionautorefname{Def.}%
    \autoref{#1}
  \endgroup
}

\definecolor{lawngreen}{rgb}{0.49, 0.99, 0.0}
\definecolor{pink}{rgb}{1, 0, 0.5}
\definecolor{airforce}{rgb}{0.36, 0.54, 0.66}

\hypersetup{
    linkbordercolor=airforce,
    urlbordercolor=pink,
    citebordercolor=green,
}

\usepackage{tikz}

\usepackage{makecell}
\AtBeginDocument{%
  }

\setcopyright{acmlicensed}
\copyrightyear{2024}
\acmYear{2024}
\acmDOI{XXXXXXX.XXXXXXX}

\acmConference[Conference acronym 'XX]{ACM Comput. Surv.}
\acmISBN{978-1-4503-XXXX-X/18/06}

\begin{document}

\title{A Survey on Diffusion Models for Time Series and Spatio-Temporal Data}

\author{Yiyuan Yang}
\affiliation{%
  \institution{University of Oxford}
  \country{UK}}
\email{yiyuan.yang@cs.ox.ac.uk}

\author{Ming Jin}
\affiliation{%
  \institution{Griffith University}
  \country{Australia}}
\email{mingjinedu@gmail.com}

\author{Haomin Wen}
\affiliation{%
  \institution{Carnegie Mellon University}
  \country{USA}}
\email{wenhaomin.whm@gmail.com}

\author{Chaoli Zhang}
\affiliation{%
  \institution{Zhejiang Normal University}
  \country{China}}
\email{chaolizcl@zjnu.edu.cn}

\author{Yuxuan Liang}
\affiliation{%
  \institution{Hong Kong University of Science and Technology (Guangzhou)}
  \country{China}}
\email{yuxliang@outlook.com}

\author{Lintao Ma}
\affiliation{%
  \institution{Ant Group}
  \country{China}}
\email{lintao.mlt@antgroup.com}

\author{Yi Wang}
\affiliation{%
  \institution{The University of Hong Kong}
  \country{China}}
\email{yiwang@eee.hku.hk}

\author{Chenghao Liu}
\affiliation{%
  \institution{Salesforce Research}
  \country{Singapore}}
\email{chenghao.liu@salesforce.com}

\author{Bin Yang}
\affiliation{%
  \institution{East China Normal University}
  \country{China}}
\email{byang@dase.ecnu.edu.cn}

\author{Zenglin Xu}
\affiliation{%
  \institution{Fudan University}
  \country{China}}
\email{zenglin@gmail.com}

\author{Shirui Pan}
\affiliation{%
  \institution{Griffith University}
  \country{Australia}}
\email{s.pan@griffith.edu.au}

\author{Qingsong Wen}
\authornote{Corresponding author}
\affiliation{%
  \institution{Squirrel Ai Learning}
  \country{USA}}
\email{qingsongedu@gmail.com}

\renewcommand{\shortauthors}{Yang et al.}

\begin{abstract}
Diffusion models have been widely used in time series and spatio-temporal data, enhancing generative, inferential, and downstream capabilities. These models are applied across diverse fields such as healthcare, recommendation, climate, energy, audio, and traffic. By separating applications for time series and spatio-temporal data, we offer a structured perspective on model category, task type, data modality, and practical application domain. This study aims to provide a solid foundation for researchers and practitioners, inspiring future innovations that tackle traditional challenges and foster novel solutions in diffusion model-based data mining tasks and applications. For more detailed information, we have open-sourced a repository\footnote{\url{https://github.com/yyysjz1997/Awesome-TimeSeries-SpatioTemporal-Diffusion-Model}}.
\end{abstract}




\maketitle

\section{Introduction} \label{sec:introduction}

Diffusion models represent a family of probabilistic generative models that undergo optimization through a two-step process involving the injection and subsequent removal of noise across a set of training samples. This process comprises a forward phase, referred to as \emph{\textbf{diffusion}}, and a reverse phase, known as \emph{\textbf{denoising}}. By training the model to remove the noise added during the diffusion process, the model learns to generate effective data samples during inference that align closely with the distribution of the training data~\cite{ho2020denoising,chang2023design}.

\begin{wrapfigure}{r}{0.34\textwidth} \vspace{-0.55cm}
    \centering
    \includegraphics[width=0.941\linewidth]{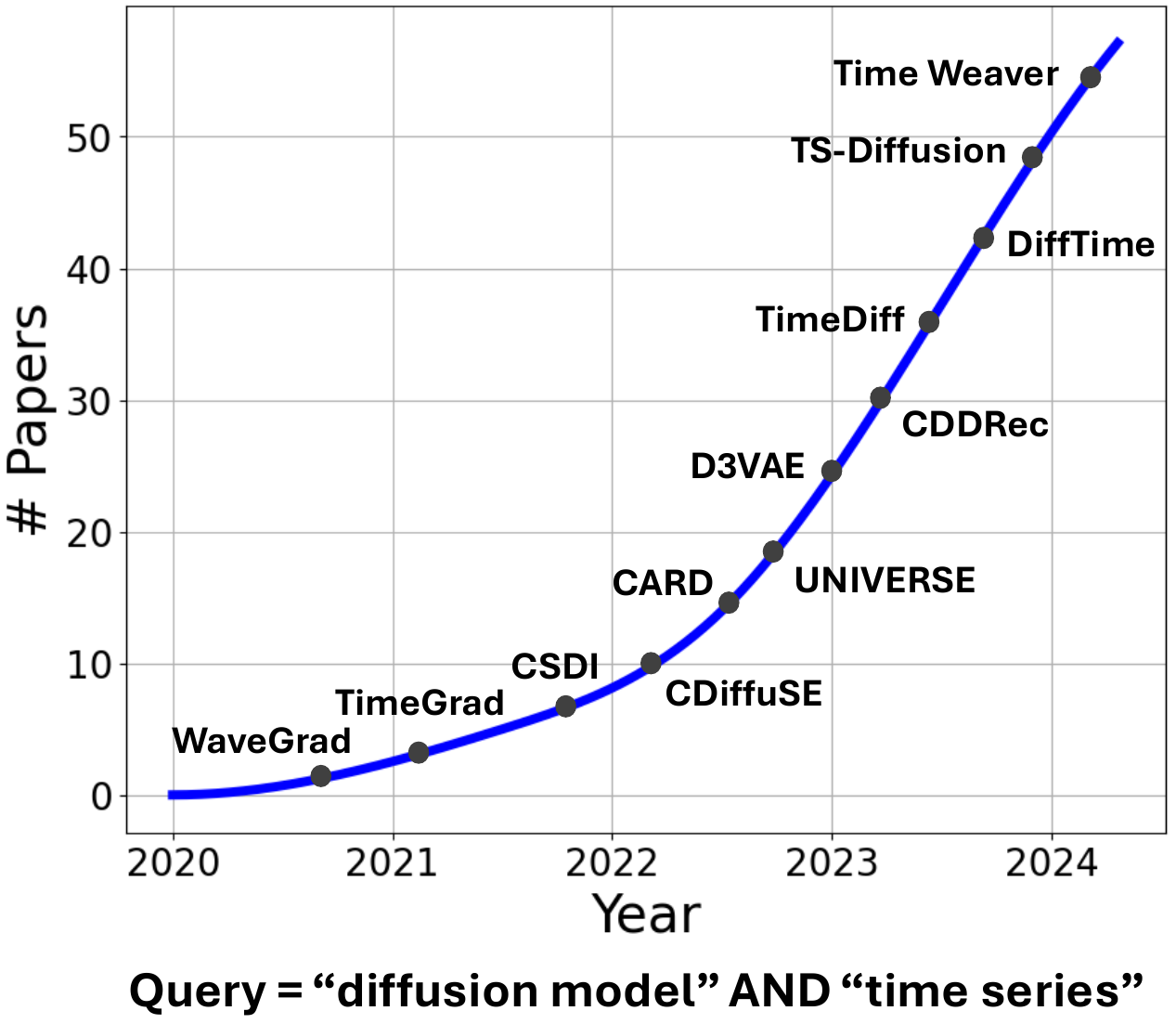}
    \vspace{0.4cm} 
    \includegraphics[width=\linewidth]{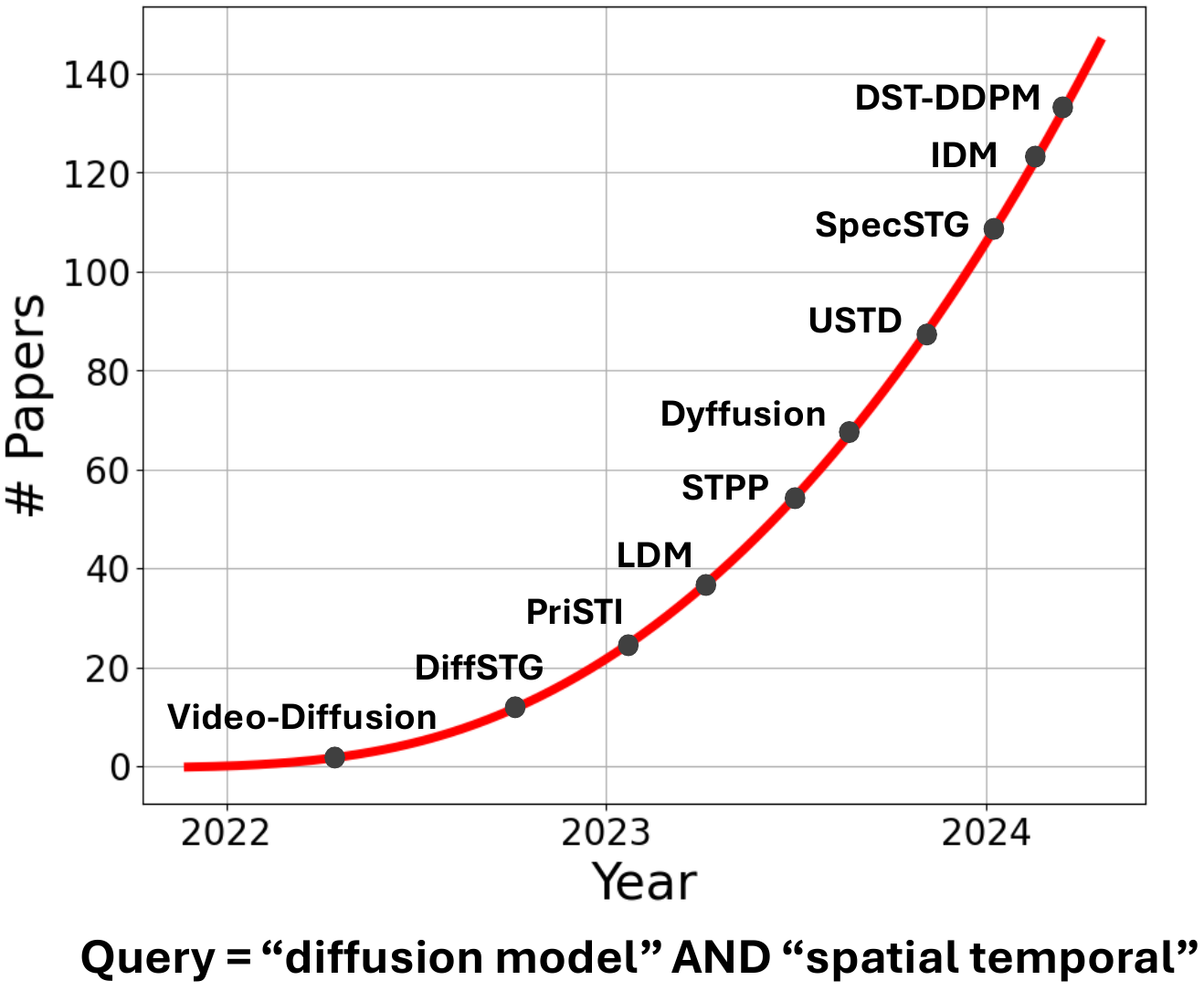} \vspace{-1.1cm}
    \caption{Trends in the cumulative number of papers related to diffusion models for time series and spatio-temporal data.} \vspace{-0.4cm}
    \label{fig:diffusion-model-trends}
\end{wrapfigure}

In recent years, diffusion models have risen to prominence and significantly influenced various domains, including computer vision (CV)~\cite{ho2020denoising,dhariwal2021diffusion,ho2022video,saharia2022palette,saharia2022image}, natural language processing (NLP)~\cite{austin2021structured,li2022diffusion,gong2022diffuseq,lovelace2024latent}, and general multimodal learning~\cite{ramesh2022hierarchical,saharia2022photorealistic,yang2023diffsound,zhang2024motiondiffuse}. This challenges the long-time supremacy of generative adversarial networks (GANs)~\cite{goodfellow2014generative,yang2023diffusionsurvey}. Within these areas, diffusion models have demonstrated remarkable capabilities in applications such as text-to-image~\cite{saharia2022photorealistic,zhang2023adding}, instance segmentation~\cite{amit2021segdiff,wolleb2022diffusion1}, 3D shape generation~\cite{vahdat2022lion,nam20223d}, molecule design~\cite{xu2022geodiff,guo2023diffusion,abramson2024accurate}, and audio generation~\cite{huang2022fastdiff,kong2020diffwave}. Remarkably, diffusion models have also gained popularity as a non-autoregressive alternative for tasks conventionally dominated by autoregressive methods~\cite{chang2023design}. Recently, the introduction of OpenAI Sora~\cite{sora2024} marked the advent of diffusion models in modeling the physical world embedded within the spacetime continuum, highlighting their critical importance. In addition, AlphaFold 3~\cite{abramson2024accurate} proposed by Google DeepMind uses diffusion models to generate 3D atomic coordinates and predict biomolecular structures like proteins, DNA, and RNA.

Temporal data, which primarily includes time series and spatio-temporal data, encapsulates the dynamics of the vast majority of real-world systems~\cite{jin2023large}. These forms of temporal data have been extensively studied and are recognized as crucial for numerous applications~\cite{jin2023survey,harutyunyan2019multitask,jin2023spatio}. However, deriving universal dynamic laws in the physical world from various data modalities remains a significant challenge within the field. Recently, the area of time series and spatio-temporal modeling has experienced a substantial shift from sensory intelligence towards general intelligence~\cite{jin2024position}. This shift is characterized by the emergence of unified foundation models (FMs) that possess versatile temporal data analytical capabilities~\cite{jin2023large,jin2024position}, challenging the supremacy of domain-specific models. Diffusion models have achieved state-of-the-art results in many modalities, including images, speech, and video~\cite{khanna2023diffusionsat}. Benefiting from the vast and diverse available data in these fields, diffusion models often serve as generative FMs alongside large language models (LLMs) or other foundation models, facilitating rapid development in these areas~\cite{rombach2022high,ho2022video}. In recent years, there has also been an increasing number of diffusion models crafted for modeling time series and spatio-temporal data (\shortautoref{fig:diffusion-model-trends}). Also, we have become aware of an increasing number of attempts to use diffusion models for temporal modeling (see \shortautoref{tab:paper_list}). Observing the success of diffusion models, an intriguing question arises: \emph{\textbf{what kind of sparks will emerge from the intersection of time series/spatio-temporal data analysis and diffusion models?}}

\begin{figure*}[!t]
    \centering
    \includegraphics[width=0.8\linewidth]{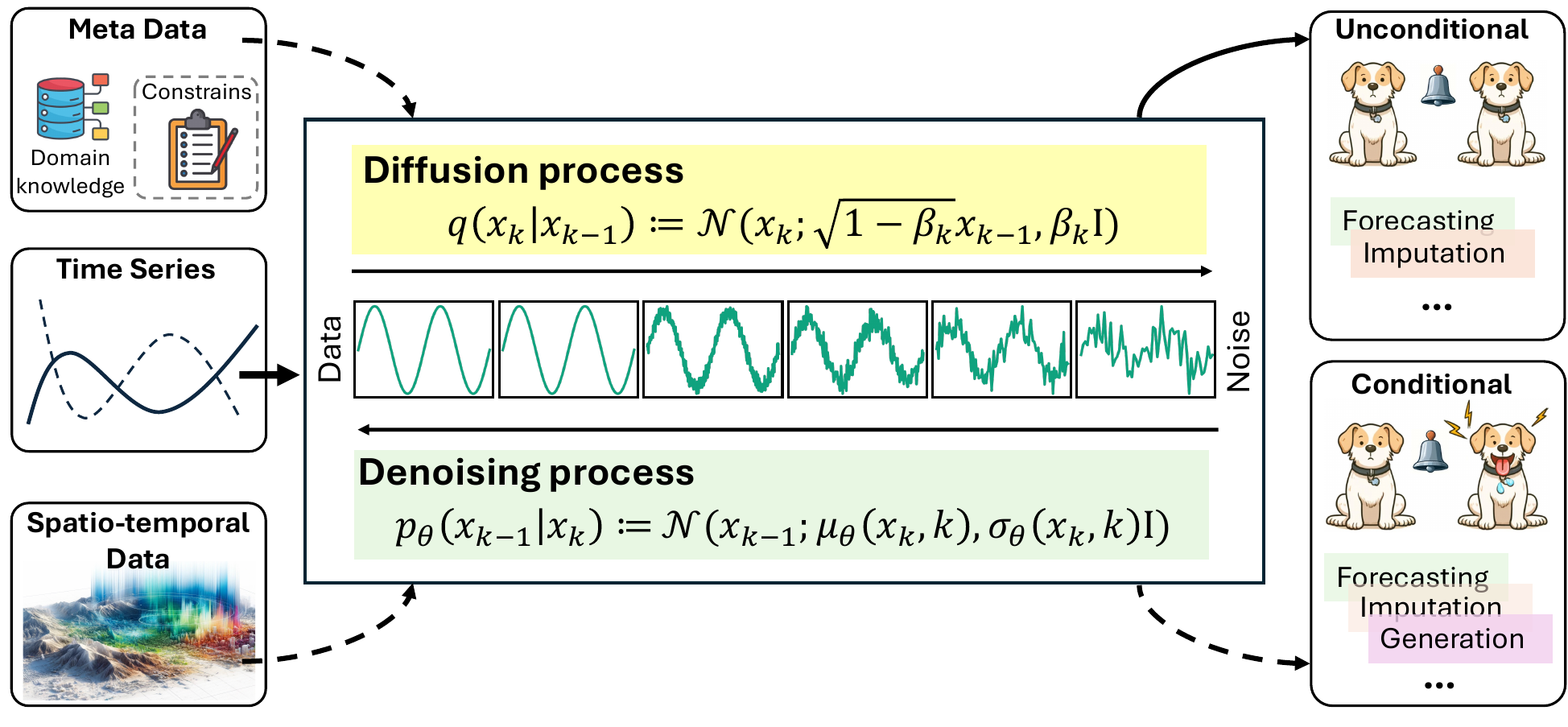} \vspace{-0.3cm}
    \caption{An overview of diffusion models for time series and spatio-temporal data analysis. In the diffusion process, $x_k$ and $x_{k-1}$ denote the results after adding noise at step $k$ and $k-1$, respectively. This process can be represented by the size of the controlling steps $\beta_{k} \in (0,1)$, the identity matrix $\mathbf{I}$, and a Gaussian distribution $\mathcal{N}(x;\mu,\sigma)$ of $x$ with the mean $\mu$ and the covariance $\sigma$. During the denoising process, the model attempts to iteratively learn the data distribution by modelling the distribution ${p}_\theta(x_{k-1}|x_k)$. The functions $\mu_\theta(\cdot)$ and variance $\sigma_\theta(\cdot)$ are the model learnable parameters.} \vspace{-0.48cm}
    \label{fig:overview}
\end{figure*}

Time series and spatio-temporal data analysis fundamentally rely on a profound understanding of their inherent temporal dynamics, wherein primary tasks predominantly focus on the generative capabilities of backbone models, such as forecasting~\cite{rasul2021autoregressive,li2022generative,feng2024latent}, imputation~\cite{tashiro2021csdi,wang2023observed,yun2023imputation,10.24963/ijcai.2025/1187} and generation~\cite{narasimhan2024time,zhang2024motiondiffuse}. These analyses center on generating temporal data samples for specific purposes in \emph{\textbf{conditional}} or \emph{\textbf{unconditional}} manners. Having witnessed the recent development of time series and spatio-temporal foundation models~\cite{jin2023large,liang2024foundation}, whether built upon LLMs or trained from scratch, their success can be attributed to the ability to estimate the distribution of training samples where effective data representations can be drawn. In this regard, diffusion models emerge as a powerful generative framework that enables (1) the modeling of complex patterns within temporal data and (2) the support of a wide range of downstream tasks, as depicted in \shortautoref{fig:overview}. 

\begin{figure*}[t]
    \centering
    \includegraphics[width=0.76\linewidth]{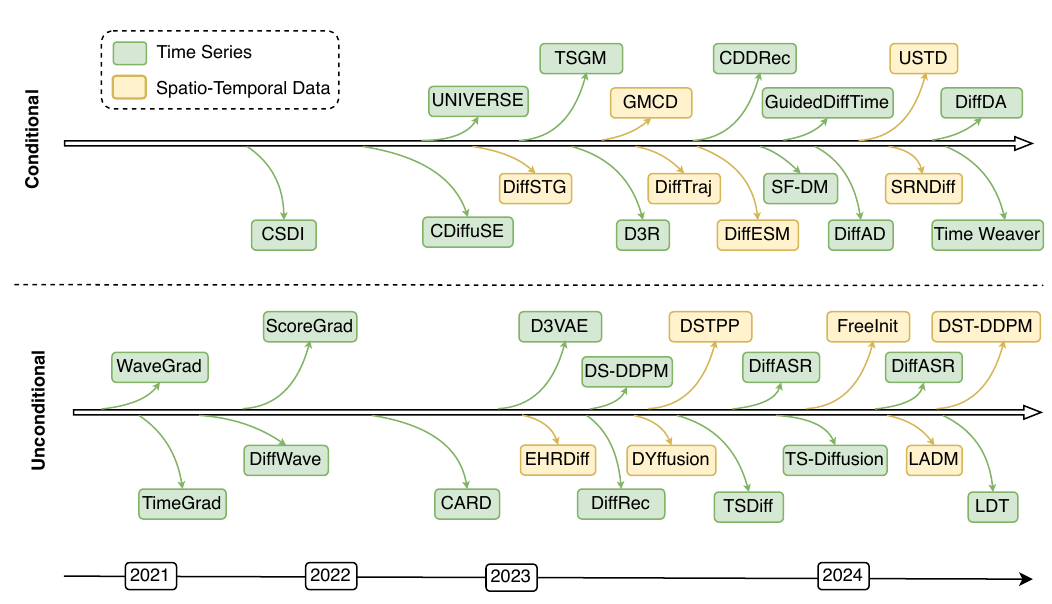}
    \caption{Representative diffusion models for time series and spatio-temporal data in recent years.} \vspace{-0.5cm}
    \label{fig:roadmap}
\end{figure*}

To generate valid data samples for specific tasks, time series and spatio-temporal diffusion models usually operate in an unconditional manner without the need for supervision signals. Given the partially-observed nature of real-world applications~\cite{wang2024timexer}, conditional diffusion models have emerged. They leverage data labels (e.g., instructions, metadata, or exogenous variables) to regulate the generation process, thereby enabling effective cross-modal prompting that leads to more tailored and improved outcomes~\cite{narasimhan2024time}. We present a roadmap in \shortautoref{fig:roadmap}. By training on large-scale temporal data, diffusion models effectively fill the gap in time series/spatio-temporal data generation and exhibit significant potential in solving the puzzle of next-generation, LLM-empowered temporal data-centric agents~\cite{wu2023next,jin2024position}.

Despite the promising prospects and rapid advancement of diffusion models in handling time series and spatio-temporal data, there has been a conspicuous lack of systematic analysis of this model family in the existing literature. This article aims to bridge this gap by providing a forward-looking review that elucidates both the 'why' and the 'how', detailing the reasons diffusion models are suited for these data modalities and unveiling the mechanisms through which they confer advantages. In this survey, we offer a detailed categorization, engage in thorough reviews, and identify burgeoning trends within this rapidly evolving landscape. Our main contributions are summarized as follows.

\begin{itemize}[leftmargin=*]
    \item \textbf{Comprehensive and up-to-date review.} We present a comprehensive, up-to-date, and forward-looking review of diffusion models for time series and spatio-temporal data. Our survey highlights the suitability of diffusion models for these data modalities and discusses the benefits they confer. By covering both a broad spectrum of the field and the specifics of individual methods, we furnish readers with a deep insight into this subject area.
    \item \textbf{Unified and structured categorization.} We introduce a clear and organized framework for categorizing the existing literature into two main types: unconditional and conditional diffusion models, focusing on time series and spatio-temporal data that span both predictive and generative tasks. This categorization offers the reader a coherent roadmap of the topic from multiple perspectives.
    \item \textbf{Insights into emerging advances.} We discuss cutting-edge techniques in both unconditional and conditional diffusion models, focusing on time series and spatio-temporal data. Our coverage includes the latest techniques and emerging trends such as multimodal conditional generation.
    \item \textbf{Summary of challenges and future directions.} We identify key challenges faced in the current research landscape and highlight several promising directions for future exploration.
\end{itemize}

The remainder of this paper is structured as follows: \shortautoref{sec:background} provides a comprehensive background on diffusion models, detailing their development, theoretical foundations, and various implementations. \shortautoref{sec:overview and categorization} presents a structured overview and categorization of diffusion models applied to time series and spatio-temporal data, setting the stage for a deeper exploration of model perspectives in \shortautoref{sec: Models}, which discusses both standard and advanced diffusion models. \shortautoref{sec: Task} focuses on task perspectives, examining how diffusion models tackle forecasting, generation, imputation, anomaly detection, and more. \shortautoref{sec: Modality} discusses data perspectives, highlighting challenges and solutions specific to time series and spatio-temporal data. \shortautoref{sec:applications} explores the application of diffusion models across various domains, such as healthcare, traffic, and energy, demonstrating their broad utility. Finally, \shortautoref{sec:future directions} concludes the paper with an outlook on future opportunities and summarizing remarks.
\section{Background} \label{sec:background}

This paper primarily reviews recent improvements in using diffusion models to solve time series and spatio-temporal data challenges. In this section, we will first define time series and spatio-temporal data, as well as their corresponding tasks in various fields. Then, we will introduce the history of the diffusion model and its advantages. Finally, some different kinds of diffusion models and their variants based on theoretical formula derivations and comparisons with other generative models will be presented.

\subsection{Overview of Time Series and Spatio-Temporal Data} 
Temporal data, particularly time series and spatio-temporal data, are important data structures for a wide range of real-world applications~\cite{jin2023large}. A time series is defined as a sequential arrangement of data points, categorized by their temporal order. These sequences may be univariate, involving a single variable over time, or multivariate, incorporating multiple variables. For instance, daily air quality measurements in a city constitute a univariate time series, while combining daily temperature and humidity readings generates a multivariate series. In our discussions, we employ bold uppercase letters (e.g., $\mathbf{X}$) to represent matrices, bold lowercase (e.g., $\mathbf{x}$) for vectors, calligraphic uppercase (e.g., $\mathcal{X}$) for sets, and standard lowercase (e.g., $x$) for scalars.

\begin{figure}[tbp]
    \centering
    \includegraphics[width=0.75\linewidth]{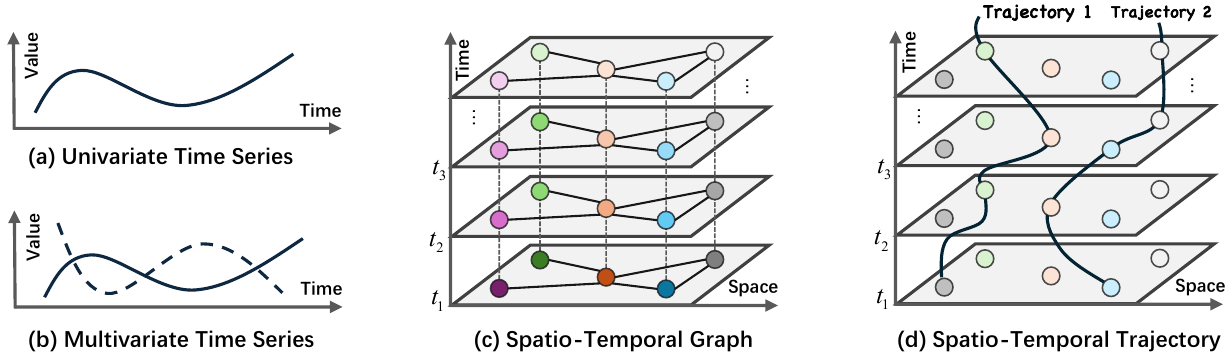}
    \caption{Illustrations of time series and spatio-temporal data.}
    \label{fig:data illustration} \vspace{-0.5cm}
\end{figure}

We base our formal definitions of time series on those provided in \cite{jin2023large,jin2023survey}. Specifically, a univariate time series (\shortautoref{fig:data illustration}(a)), denoted as $\mathbf{x} = (x_1, x_2, \ldots, x_T) \in \mathbb{R}^{T}$, consists of a sequence of $T$ data points arranged chronologically, where each $x_t \in \mathbb{R}$ represents the series' value at time $t$. Conversely, a multivariate time series (\shortautoref{fig:data illustration}(b)), represented by $\mathbf{X} = (\mathbf{x}_1, \mathbf{x}_2, \ldots, \mathbf{x}_T) \in \mathbb{R}^{T \times D}$, encompasses a sequence of $T$ data points also in temporal sequence but across $D$ feature channels, with $\mathbf{x}_t \in \mathbb{R}^{D} (1 \leq t \leq T)$ indicating the series' values at time $t$ across $D$ different channels. For a comprehensive exploration of time series, we direct the reader to \cite{jin2023survey}.

Spatio-temporal data, in contrast, encompasses sequences of data points characterized by both their temporal and spatial dimensions. This type of data integrates the aspect of time, as seen in time series, with the additional dimension of space, capturing the complex dynamics of phenomena as they unfold over time and across different locations. From this perspective, multivariate time series may also be considered as a form of spatio-temporal data. Such data is instrumental in fields ranging from geography and meteorology to urban planning and environmental monitoring, where understanding the interplay between spatial patterns and temporal evolution is crucial.

In practical applications, spatio-temporal data refers to a collection of observations where each data point is defined by its position in space and time, encapsulating a diverse range of data structures such as graphs, trajectories, and even videos, as noted in \cite{jin2023large}. 

For instance, a spatio-temporal graph (\shortautoref{fig:data illustration}(c)) representing urban traffic flow \cite{sun2020predicting,liang2021fine} over time can be understood as spatio-temporal data where each node represents a specific location with specific attributes and edges are weighted by traffic volume, which changes over time. Likewise, trajectory data \cite{liang2021modeling,chen2024deep} (\shortautoref{fig:data illustration}(d)) captures the movement of objects through space over time, including their paths, speeds, and changes in direction. Such data is vital for applications in transportation studies, wildlife tracking, and mobile network optimization, where analyzing the patterns of movement and predicting future locations based on historical data are of paramount importance. On this basis, these and other spatio-temporal constructs can be systematically characterized and analyzed.

Building on the definitions provided, we now proceed to succinctly introduce the representative tasks associated with each data category~\cite{jin2023large}.

\begin{itemize}[leftmargin=*]
\item \textbf{Time Series Analysis.} The analysis of time series with diffusion models encompasses four primary tasks: forecasting, generation, anomaly detection, and imputation. Forecasting focuses on predicting future values within a time series, which can be subdivided into short-term and long-term forecasts based on the temporal scope of the predictions. Generation involves creating new time series based on the statistical properties of a given dataset, serving as a way to simulate possible scenarios or enhance data diversity for training models. Anomaly detection, a specialized form of classification, aims to distinguish atypical series from normal ones. Imputation addresses the challenge of filling in missing values within a series, which is crucial for maintaining the integrity and utility of time series.
\item \textbf{Spatio-Temporal Data Analysis.} Spatio-temporal data analysis, while encompassing tasks similar to those in time series analysis, often applies these methodologies within specific application scenarios. For instance, forecasting may focus on traffic flow~\cite{wen2023diffstg} or air quality~\cite{chen2024quantifying}, utilizing historical data patterns to predict future conditions. Generation tasks might involve creating synthetic trajectories, offering privacy-compliant alternatives to original datasets by replacing sensitive information with generated, anonymized data~\cite{zhu2024difftraj,zhu2024controltraj}. Anomaly detection becomes particularly crucial in scenarios such as vehicle trajectory analysis, where deviations from generated normative patterns may indicate unusual or suspicious behaviors~\cite{li2024difftad}. Additionally, spatio-temporal imputation plays a vital role in addressing missing values in multivariate time series~\cite{yun2023imputation,du2024tsi}, ensuring comprehensive and accurate datasets for further analysis. More discussion is given in \shortautoref{sec: Task}.
\end{itemize}

\vspace{-0.3cm}
\subsection{Why Diffusion Model and Its History}
Diffusion models are a class of probability-based generative models. They are called after the mathematical process of diffusion, which is commonly used to describe phenomena such as particle movement in a gas or liquid~\cite{sohl2015deep}. In detail, the concept of diffusion models first appeared in statistical physics, used to describe the process of particles moving from areas of high concentration to areas of low concentration~\cite{jarzynski1997equilibrium}. Early diffusion models were primarily concerned with accurately simulating the random diffusion behavior in the generation process. 

One of the key breakthroughs in diffusion models occurred in 2015, when researchers proposed a method that combines variational inference to effectively train these models~\cite{sohl2015deep}. Since then, the field has experienced rapid development, especially in the area of high-resolution image generation~\cite{rombach2022high}. Since 2020, diffusion models have begun to show their potential in more fields, such as text2image, music generation, and speech synthesis~\cite{saharia2022photorealistic, wang2023diffuseroll, richter2023speech}. These advances are due to the optimization of model structures, improvements in training methods, and increased computational resources.

In addition to the field of application, in terms of theory, researchers began to explore how to generate data with specific features by controlling the reverse process. This study direction finally resulted in the development of diffusion models capable of producing high-quality, multidimensional data. In updated computer algorithms, it can be represented as the process of gradually modifying the data distribution, progressively injecting noise until it matches the target distribution, in order to obtain high-quality and realistic synthetic data samples~\cite{ho2020denoising}. 

Recently, more and more researchers and engineers are now focusing their perspectives on the diffusion model, and it has become one of the first choices for generating models~\cite{yang2023diffusionsurvey}. Diffusion models excel at generating high-quality, complex sequences, including time series and spatio-temporal data, with detailed coherence by gradually removing noise. They offer strong control over generations, allowing fine-tuning based on the conditions~\cite{croitoru2023diffusion}. These models are flexible and adaptable across various data types and modalities, robust against errors with a gradual noise reduction mechanism, and capable of exploring data diversity for creative outputs. Moreover, they can be integrated with other model types, like autoencoders, to enhance generation quality and control~\cite{yang2023diffusionsurvey}.

\subsection{Typical Diffusion Models} 

Typically, the training process includes two steps: the \emph{\textbf{forward process (diffusion)}} and the \emph{\textbf{reverse process (denoising)}}. Diffusion models start with a noise distribution, which is gradually altered through a series of steps. In the forward process, the model incrementally adds noise to the original data over multiple steps until the data turns into pure random noise. This process is usually Markovian, meaning that each step depends only on the preceding one. Then the reverse process takes place, involving learning to remove the noise from the data, essentially reversing the forward process. By training the model to remove the noise added during the diffusion process, the model learns to generate samples from the same distribution as the training data. The entire training process involves optimizing the model to denoise effectively. This is typically done using a loss function that encourages the model to produce samples that are close to the true data distribution~\cite{nichol2021improved}.

The currently common frameworks for diffusion models include denoised diffusion probabilistic models (DDPMs)~\cite{sohl2015deep,ho2020denoising}, score-based stochastic differential equations (Score SDEs)~\cite{song2020score,song2019generative}, conditional diffusion models~\cite{rombach2022high,dhariwal2021diffusion,ho2022classifier}, etc. Following, we will introduce the subclasses of the diffusion model through theoretical formula derivation.

\subsubsection{Denoised Diffusion Probabilistic Models (DDPMs)} \label{subsubsec:DDPM}
Denoised diffusion probabilistic models are built around a well-defined probabilistic process via dual Markov chains that consist of two parts: a diffusion (or forward) process that gradually transforms the data into noise with pre-determined noise, such as Gaussian noise, and a denoising (or reverse) process that attempts to recover the original data by deep neural networks. 

\textbf{Forward (Diffusion) Process}. Given a data distribution $q(\mathbf{x})$ and sample an initial clean data $\mathbf{x}_0 \sim q(\mathbf{x}_0)$ from it. The subsequent forward diffusion process incrementally adulterates the initial data distribution by superimposing Gaussian noise, and finally progresses towards convergence with the standard Gaussian distribution. In the diffusion process up to step $K$, a sequence of distributed latent data $\mathbf{x}_1,\mathbf{x}_2,\cdots,\mathbf{x}_K$ materializes. The diffusion process can be defined as a Markov chain transforms $\mathbf{x}_{k-1}$ to $\mathbf{x}_k$ with a diffusion transition kernel:

\begin{equation} \label{eq:ddpm-forward1}
    q(\mathbf{x}_k | \mathbf{x}_{k-1}) := \mathcal{N}(\mathbf{x}_{k} ; \sqrt{1-\beta_{k}} \mathbf{x}_{k-1}, \beta_{k} \mathbf{I}), 
\end{equation}

\noindent for $\forall {k} \in \{ 1, \cdots, K \}$ with the size of the controlling steps $\beta_{k} \in (0,1)$, the identity matrix $\mathbf{I}$, and a Gaussian distribution $\mathcal{N}(\mathbf{x};\mu,\sigma)$ of $\mathbf{x}$ with the mean $\mu$ and the covariance $\sigma$. According to the properties of the Gaussian kernel, it is feasible to get $\mathbf{x}_k$ directly from $\mathbf{x}_0$ by equation~\ref{eq:ddpm-forward1},  and collect noise samples straight from the original input $\mathbf{x}_0$ for any step, that is,

\begin{equation} \label{eq:ddpm-forward2}
    q(\mathbf{x}_{k} | \mathbf{x}_{0}) := \prod_{k=1}^K {q}(\mathbf{x}_{k} | \mathbf{x}_{k-1}) := \mathcal{N}(\mathbf{x}_{k} ; \sqrt{\bar{\alpha_{k}}} \mathbf{x}_0, \sqrt{1 - \bar{\alpha_{k}}} \mathbf{I}),
\end{equation}

\noindent with $\alpha_{k} := 1 - \beta_{k}$ and $\bar{\alpha_{k}} := \prod_{i=1}^{K} \alpha_{i}$. Therefore, $\mathbf{x}_{k} = \sqrt{\bar{\alpha_{k}}}  \mathbf{x}_0 + \sqrt{1 - \bar{\alpha_{k}}} \epsilon$ with Gaussian noise $\epsilon \sim \mathcal{N}(0,\mathbf{I})$. Typically, it is designed $\bar{\alpha_{k}} \approx 0$, s.t., $q(\mathbf{x}_{k}) := \int q( \mathbf{x}_{K} | \mathbf{x}_{0} )q(\mathbf{x}_{0})\mathbf{d}\mathbf{x}_{0} \approx \mathcal{N}(\mathbf{x}_{k};\mathbf{0},\mathbf{I})$, i.e., the backward chain can begin with any Gaussian noise. Overall, the forward process gradually injects noise into the data until all structures have disappeared.

\textbf{Reverse (Denoising) Process}. The reverse process performs the denoising task at each step with a series of Markov chains until the damaged original data is reconstructed. Specifically, the series of reverse Markov chains start with a distribution $p(\mathbf{x}_K)=\mathcal{N}(\mathbf{x}_K;\mathbf{0},\mathbf{I})$ and a learnable kernel $p_\theta(\mathbf{x}_{k-1} | \mathbf{x}_k)$ to generate $p_\theta(\mathbf{x}_0)$. The learnable Gaussian transition kernels $p_\theta$ can be represented as:

\begin{equation} \label{eq:ddpm-reverse1}
    {p}_\theta (\mathbf{x}_{k-1} | \mathbf{x}_{k}) := \mathcal{N}(\mathbf{x}_{k-1} ; \mu_\theta(\mathbf{x}_{k}, {k}), \sigma_\theta(\mathbf{x}_{k}, {k}) \mathbf{I}), 
\end{equation}

\noindent where the mean $\mu_\theta(\cdot)$ and variance $\sigma_\theta(\cdot)$ are the model learnable parameters. The model tries to learn the data distribution by the model distribution ${p}_\theta(\mathbf{x}_0)$ during the reverse denoising process.

\textbf{Training}. In order to approximate the real data distribution, the diffusion model is trained to minimize variational constraints on the negative log-likelihood (NLL):

\begin{equation} \label{eq:ddpm-training1}
\begin{aligned}
\mathbb{E}\left[-\log p_{\theta}\left(\mathrm{x}_{0}\right)\right] &\leq \mathbb{E}_{q}\left[-\log \frac{p_{\theta}\left(\mathrm{x}_{0: K}\right)}{q\left(\mathrm{x}_{1: K} \mid \mathrm{x}_{0}\right)}\right]  =\mathbb{E}_{q}\left[-\log p\left(\mathrm{x}_{K}\right)-\sum_{k \geq 1} \log \frac{p_{\theta}\left(\mathrm{x}_{k-1} \mid \mathrm{x}_{k}\right)}{q\left(\mathrm{x}_{k} \mid \mathrm{x}_{k-1}\right)}\right] =: L.
\end{aligned}
\end{equation}

\noindent It is equivalent to Kullback–Leibler divergence (KL divergence) format as mentioned in the DDPM paper~\cite{ho2020denoising} with three parts: the prior loss $L_K$,  the divergence of the forwarding step and the corresponding reversing step $L_{k-1}$, and the reconstruction loss $L_0$:

\begin{equation} \label{eq:ddpm-training2}
\begin{aligned}
L := & \mathbb{E}_{q}[\underbrace{D_{\mathrm{KL}}\left(q\left(\mathrm{x}_{K} \mid \mathrm{x}_{0}\right) \| p\left(\mathrm{x}_{K}\right)\right)}_{L_{K}} +\sum_{k>1} \underbrace{D_{\mathrm{KL}}\left(q\left(\mathrm{x}_{k-1} \mid \mathrm{x}_{k}, \mathrm{x}_{0}\right) \| p_{\theta}\left(\mathrm{x}_{k-1} \mid \mathrm{x}_{k}\right)\right)}_{L_{k-1}} - \underbrace{\log p_{\theta}\left(\mathrm{x}_{0} \mid \mathrm{x}_{1}\right)}_{L_{0}}].
\end{aligned}
\end{equation}

\noindent Especially, to minimize the NLL, we can only train the divergence loss between two steps $L_{k-1}$, and using Baye's rule to parameterize the posterior $q\left(\mathrm{x}_{k-1} \mid \mathrm{x}_{k}, \mathrm{x}_{0}\right)$, that is:

\begin{equation} \label{eq:ddpm-training3}
q\left(\mathrm{x}_{k-1} \mid \mathrm{x}_{k}, \mathrm{x}_{0}\right)=\mathcal{N}\left(\mathrm{x}_{k-1} ; \tilde{\boldsymbol{\mu}}_{k}\left(\mathrm{x}_{k}, \mathrm{x}_{0}\right), \tilde{\beta}_{k} {I}\right),
\end{equation}

\begin{equation} \label{eq:ddpm-training4}
\tilde\mu_t(\mathbf{x}_k, \mathbf{x}_0) := \frac{\sqrt{\bar\alpha_{k-1}}\beta_k }{1-\bar\alpha_k}\mathbf{x}_0 + \frac{\sqrt{\alpha_k}(1- \bar\alpha_{k-1})}{1-\bar\alpha_k} \mathbf{x}_k \hspace*{1mm},
\end{equation}

\begin{equation} \label{eq:ddpm-training5}
\tilde\beta_k := \frac{1-\bar\alpha_{k-1}}{1-\bar\alpha_k}\beta_k.
\end{equation}

\noindent where $\alpha_k$ is $1-\beta_k$ and $\bar{\alpha}_k$ indicates $\prod_{k=1}^{K} \alpha_{k}$. $L_{k-1}$ can be equated to the expected value of the $\ell_2$-loss between the two mean coefficients:

\begin{equation}
L_{k-1}=\mathbb{E}_{q}\left[\frac{1}{2 \sigma_{k}^{2}}\left\|\tilde{\boldsymbol{\mu}}_{k}\left(\mathrm{x}_{k}, \mathrm{x}_{0}\right)-\boldsymbol{\mu}_{\theta}\left(\mathrm{x}_{k}, k\right)\right\|^{2}\right]+C.
\end{equation}

\noindent Ho et al.~\cite{ho2020denoising} emphasize that, rather than parameterizing the mean $\mu_\theta(\mathrm{x}_k,k)$, predicting the noise vector at each time step in the forward process by parameterizing $\epsilon_\theta(\mathrm{x}_k,k)$ for simplification:

\begin{equation} \label{eq:ddpm-training6}
    \mathbb{E}_{k \sim \mathcal{U} (1,K), \mathbf x_0 \sim q(\mathbf x_0), \epsilon \sim \mathcal{N}(\mathbf{0},\mathbf{I})} \bigg[ {\lambda(k)  \left\| \epsilon - \epsilon_\theta(\mathbf{x}_k, k) \right\|^2} \bigg],
\end{equation}

\noindent where $\lambda(k) = \frac{{\beta_k}^2}{2{\sigma_K}^2\alpha_k(1-\bar{\alpha}_k)}$ is a weight that changes the noise scale and $\epsilon_\theta$ is a model for Gaussian noise prediction. After training using the above loss function, $\epsilon_\theta$ will be used in the reverse process of ancestor sampling.

\textbf{Inference (Sampling)}. Given the noisy data $\mathrm{x}_K$ and starting with step $K$ denoising, the final time series is generated through the equation:

\begin{equation} \label{eq:ddpm-inference1}
\begin{aligned} p_\theta(\mathbf{x}_{k-1}|\mathbf{x}_k) &= \mathcal{N}(\mathbf{x}_{k-1}; \mu_\theta(\mathbf{x}_k, k), \sigma_\theta^2(\mathbf{x}_k, k)\mathbf{I}) \sim \frac{1}{\sqrt{\alpha_k}}(\mathbf{x}_k - \frac{\beta_k}{\sqrt{1-\overline{\alpha_k}}}\epsilon_\theta(\mathbf{x}_k, k)) + \sigma_\theta(\mathbf{x}_k, k)z,
\end{aligned} \end{equation}

\noindent where $z \sim \mathcal{N}(0, \mathbf{I})$, also $\beta_k \approx \sigma_\theta^2(\mathbf{x}_k, k)$ in practice.

\subsubsection{Score SDE Formulation} \label{subsubsec:SDE}
DDPM achieved a set of discrete steps in the forward processing, so it has some limitations about training designs. Score SDE further generalizes DDPM's discrete system to a continuous framework based on the stochastical differential equation~\cite{song2020score}. Here we use $T$ instead of the step size $k$ in DDPM. 

\textbf{Forward Process.} The corresponding continuous diffusion process can be represented using Itô SDE~\cite{ito1951stochastic}, including a mean shift and a Brownian motion (standard Wiener process) as follows:

\begin{equation} \label{eq:sde-forward}
	\mathrm{d} \mathbf{x}=\mathbf{f}(\mathbf{x}, t) \mathrm{d} t+g(t) \mathrm{d} \mathbf{w}, t \in [0, T],
\end{equation}

\noindent where $\mathbf{f}(\cdot, t)$ represents the drift coefficient for the stochastic process $\mathbf{x}(t)$, and $g(\cdot)$ is the diffusion coefficient linked to the Brownian motion $\mathbf{w}$.

\textbf{Reverse Process.} The new samples can be synthesized from the known prior distribution $p_T$ by solving the reverse-time SDE~\cite{anderson1982reverse}:

\begin{equation} \label{eq:sde-reverse}
	\mathrm{d} \mathbf{x}=\left[\mathbf{f}(\mathbf{x}, t)-g^2(t) \nabla_\mathbf{x} \log p_t(\mathbf{x}) \right] \mathrm{d} t+g(t) \mathrm{d} \bar{\mathbf{w}},
\end{equation}

\noindent where $\bar{\mathbf{w}}$ is a Brownian motion with reversed time flows~\cite{vincent2011connection}. The solution to the reverse-time SDE is approximated by a time-dependent neural network $s_\theta(\mathbf{x}, t)$ to a score function $\nabla_\mathbf{x} \log p_t{(\mathbf{x})}$. The solution of the forward SDE equation is that:

\begin{equation}
\begin{aligned}
{L} :=  \mathbb{E}_{t}&\{\lambda(t) \mathbb{E}_{\mathbf{x}_0}  \mathbb{E}_{q(\mathbf{x}_t| \mathbf{x}_0)}[\|{s}_{\boldsymbol{\theta}}(\mathbf{x}_t t)-\nabla_{\mathbf{x}_t} \log p(\mathbf{x}_t| \mathbf{x}_0)\|_{2}^{2}]\}, 
\label{eq:sde-result}
\end{aligned}
\end{equation}

\noindent where $\mathbf{x}_0$ is sampled from the distribution $p_0$ and $\lambda(t)$ is the
positive weighting function. This method circumvents the direct approximation of the computationally infeasible score function by estimating the transition probability that follows a Gaussian distribution during the forward diffusion process~\cite{song2020score}.

Besides, there are some simplified explanations of reverse-time SDE solvers. Using these techniques, samples can be generated after training using various methods. Euler-Maruyama (EM) Method~\cite{mao2015truncated}: Solves the reverse-time SDE through a simple discretization technique, replacing $\mathrm{d} \mathbf{x}$ with $\triangle \mathbf{t}$ and $\mathrm{d} \bar{\mathbf{w}}$ with Gaussian noise $z$. Prediction-Correction (PC) Method: Operates in a sequential manner, alternating between predictor and corrector steps. The predictor can use any numerical solver, such as the EM method, for the reverse-time SDE, while the corrector can be any score-based Markov Chain Monte Carlo (MCMC) method. Probability Flow ODE Method~\cite{song2020score}: Reformulates the forward SDE into an ODE that maintains the same marginal probability density $p_t$ as the SDE. Sampling by solving this reverse-time ODE is equivalent to solving the time-reversed SDE. There are also some advanced ODE solvers to speed up the process~\cite{liu2022pseudo,lu2022dpm}.


\vspace{-0.2cm}
\subsubsection{Conditional Diffusion Models} \label{CDM}
In the previous sections, we discussed the DDPM and Score SDE from an unconditional view. They generate data without any explicit conditions or guidance. The model learns to produce outputs from the learned distribution of the input data. The general diffusion models are capable of generating data samples not just from an unconditional distribution $p_0$, but also from a conditional distribution $p_0(\mathbf{x}|c)$ when given a condition $c$. This condition could be class labels or features related to the input data $\mathbf{x}$~\cite{rombach2022high}. During training, the score network $s_\theta(\mathbf{x}, t, c)$ takes condition $c$ as an input. Additionally, there are specific sampling algorithms designed for conditional generation, such as label-based conditions~\cite{dhariwal2021diffusion}, label-free conditions~\cite{ho2022classifier}, and further distillation-based guidance~\cite{dhariwal2021diffusion}, self guidance~\cite{epstein2024diffusionself,yang2022score}, textual-based guidance~\cite{le2024voicebox,gong2022diffuseq}, graph-based guidance~\cite{schneuing2022structure}, physical-based guidance~\cite{yuan2023physdiff}, task-based guidance~\cite{an2023diffusionnag}, etc. These conditional mechanisms are more conducive to the generation of application-specific fields by using the control of other information to generate results~\cite{yang2023diffusionsurvey}.

In detail, sampling under labels and classifiers' conditions involves using gradient guidance at each step, which typically requires an additional classifier with encoder architecture (e.g., U-Net~\cite{ronneberger2015u} and Transformer~\cite{vaswani2017attention}) to generate condition gradients for specific labels~\cite{dhariwal2021diffusion}. These labels are flexible and can be textual or categorical, binary, or based on extracted features~\cite{dhariwal2021diffusion,lu2022dpm,nichol2021glide,meng2023distillation,hu2022global,wolleb2022diffusion1,chen2023diffusiondet,baranchuk2021label,yang2024generate,zuo2023unsupervised}. Correspondingly, sampling under unlabeled conditions relies solely on self-information for guidance~\cite{epstein2024diffusionself,chao2022quasi,kollovieh2024predict}. Compared to the high accuracy of the labeled conditional diffusion model, the unlabeled one has advantages in generating innovative and diverse data. Therefore, unlabelled models are more suitable for exploratory and creative application scenarios~\cite{chung2022come,choi2021ilvr}. Furthermore, there are also more conditional methods being proposed, which use information about the data itself~\cite{kollovieh2024predict}, other modalities~\cite{nichol2021glide,liu2023generating}, other representations~\cite{yuan2023physdiff}, and other knowledge~\cite{shao2024data} as conditions to guide the diffusion model for generation. Currently, the condition-based diffusion model is also the most common method in various application scenarios due to its highly specific and controlled outputs~\cite{bansal2023universal,zhang2023adding}.

\vspace{-0.2cm}
\subsubsection{Improvements with Diffusion Model and Its Variants} \label{DM-Variants}

While diffusion models have yielded satisfactory results in various tasks, practical applications reveal limitations such as slow iterative sampling and computational complexity due to high-dimensional input. There are also concerns regarding generalization performance and integration with other generative models. Next, we will explore the variants of diffusion models from efficiency and performance improvement, highlighting the modifications and optimizations they bring in comparison to the original diffusion models.

\textbf{Efficiency Enhancement}. (1) {Forward Processes Improvement}:
Several innovations aim to enhance model efficiency and robustness by using physical phenomena. PFGM~\cite{Xu2022PoissonFG} and its extension PFGM++~\cite{Xu2023PFGMUT} and \cite{Liu2023GenPhysFP} use electric field dynamics and augmented dimensions for improved performance. Cold Diffusion uses image transformations as a forward process~\cite{bansal2022cold}, and improved Gaussian perturbation kernel methods were presented~\cite{karras2022elucidating}. (2) {Reverse Processes Improvement}: Methods to reduce generation steps or use lightweight models include training-free sampling methods that optimize the trajectory from noise to data distributions, such as DDIM~\cite{song2020denoising}, gDDIM~\cite{zhang2022gddim}, PNDM~\cite{liu2022pseudo}, EDM~\cite{karras2022elucidating}, DEIS~\cite{zhang2022fast}, and DPM-Solver~\cite{lu2022dpm}. SDE-based methods use techniques like restart sampling~\cite{jolicoeur2021gotta, xu2023restart}. Knowledge distillation is also used by transferring insights from larger models to simpler ones~\cite{berthelot2023tract, song2023consistency}. (3) {Integration with Other Models}: Combining diffusion models with VAE~\cite{lyu2022accelerating, pandey2022diffusevae} and GAN~\cite{zheng2022truncated, watson2021learning} improves efficiency. Using latent space as input can reduce dimensionality and enhance efficiency~\cite{pandey2022diffusevae, rombach2022high, vahdat2021score, zhang2023dimensionality}. (4) {Scheduler Functions}: Optimizing the reverse process with improved schedulers such as CMS~\cite{song2023consistency}, DDIM~\cite{song2020denoising}, IDDPM~\cite{nichol2021improved}, DEIS~\cite{zhang2022fast}, DPM-Solvers~\cite{lu2022dpm, lu2022dpm2}, Euler and Heun scheduler~\cite{karras2022elucidating}, LCM~\cite{luo2023latent}, RePaintScheduler~\cite{lugmayr2022repaint}, TCD~\cite{zheng2024trajectory}, UniPC~\cite{zhao2024unipc}, and VQD~\cite{gu2022vector} enables faster convergence and reduces iterations.

\textbf{Performance Enhancement}. (1) {Model Architecture}: Traditional DDPM uses the U-Net architecture for efficiency. Innovations include normalization~\cite{kawar2022denoising, berthelot2023tract} and attention mechanisms with position encoding~\cite{zhang2022fast, rissanen2022generative, kingma2021variational}. Transformer-based architectures are increasingly used to improve performance~\cite{vaswani2017attention}, including ViT~\cite{dosovitskiy2020image}, SwinTransformer~\cite{liu2021swin, liu2022swin}, DiT~\cite{peebles2023scalable}, and SiT~\cite{ma2024sit}. (2) {Training Schedules}: Advanced decoding strategies involve optimizing diffusion stages and innovative projection methods. Optimization strategies adjust the diffusion steps~\cite{zheng2022truncated, lyu2022accelerating}, while projection techniques use various diffusion processes to increase versatility~\cite{hoogeboom2022blurring, daras2022soft}. Innovations in noise adjustment~\cite{kingma2021variational, kong2021fast, nichol2021improved} and optimization methods focusing on loss and matching~\cite{song2021maximum, huang2021variational} enhance model performance. (3) {Data Bridging}: To address generating arbitrary Gaussian distributions with complex distributions, research uses SDE/ODE principles. Techniques like $\alpha$-blending~\cite{heitz2023iterative} and Rectified Flow~\cite{liu2022flow, albergo2022building} explore ODE creation between distributions. The Schrödinger Bridge concept offers new avenues in distribution transportation~\cite{su2022dual, liu20232}. (4) {Conditional and Guidance Strategies}: These strategies are crucial for directing generation and enhancing output relevance and quality in response to specific conditions, as discussed in Sec~\ref{CDM}.

Currently, institutions like OpenAI and Stability AI have launched numerous outstanding diffusion models. For instance, Stable Diffusion generates high-quality images from textual descriptions~\cite{rombach2022high}, and ControlNet enhances control over generated images~\cite{zhang2023adding}. The consistency model offers state-of-the-art performance without the slow iterative process of traditional models~\cite{song2023consistency}.

\vspace{-0.3cm}
\subsection{Diffusion Model vs Other Generative Models} 

\begin{figure}
    \centering
    \includegraphics[width=0.8\linewidth]{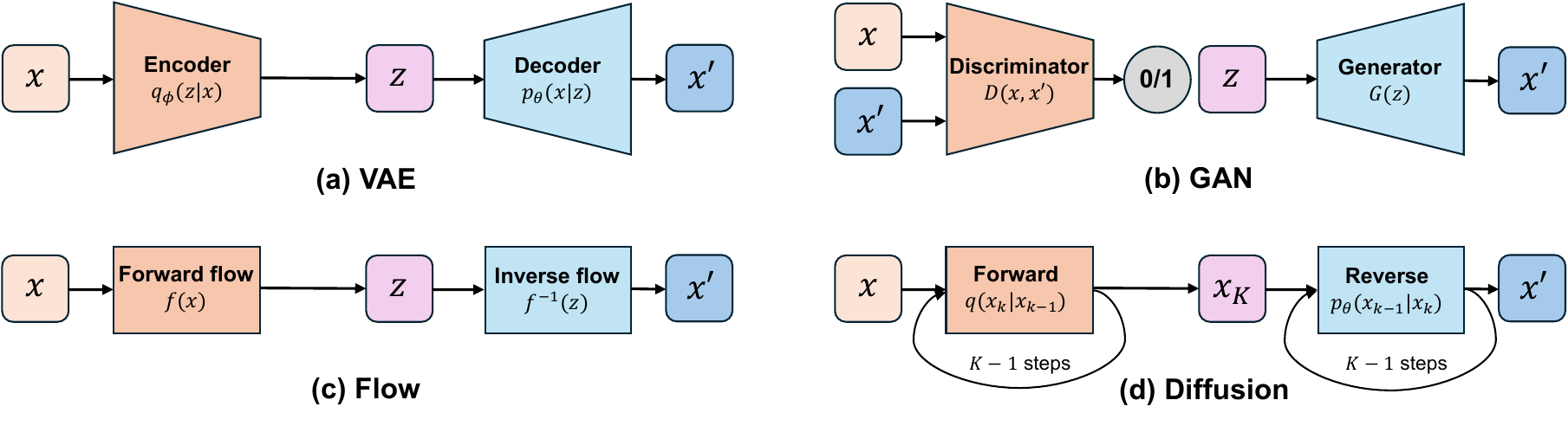} \vspace{-0.3cm}
    \caption{Different types of generative models.} \vspace{-0.5cm}
    \label{fig:generative models}
\end{figure}

In addition to the diffusion model, there are many classical generative models. We take the most widely used variational autoencoders (VAEs), generative adversarial networks (GANs), and flow-based generative models as examples to introduce them and analyse their advantages, disadvantages and differences with the diffusion model. The workflow of these models is shown in~\shortautoref{fig:generative models}.

VAE is a probabilistic generative model that encodes input data into a latent space $z$ and decodes from that latent space to generate data~\cite{kingma2013auto}, as shown in~\shortautoref{fig:generative models}(a). VAEs aim to maximize the lower bound of the log-likelihood of the data, known as the Evidence Lower Bound (ELBO). GAN consists of a generator $G(z)$ and a discriminator $D(x,x')$, as shown in~\shortautoref{fig:generative models}(b). The generator tries to produce samples as close to real data $x$ as possible, while the discriminator tries to distinguish between real data $x$ and generated data $x'$. The training of GANs involves a zero-sum game, continually optimizing both the generator and the discriminator~\cite{goodfellow2014generative}. Flow-based generative model (normalizing flows) transform data into a simpler distribution (e.g., Gaussian) through a sequence of reversible transformations, as shown in~\shortautoref{fig:generative models}(c), ensuring precise likelihood evaluation due to the transformations being invertible~\cite{rezende2015variational}.

Different generative models have their unique strengths and limitations, making them suitable for specific applications. VAE is appreciated for its simplicity, stability, and clear theoretical foundation, but tends to produce lower-quality data and has limited expressiveness in its latent spaces. GAN is known for its powerful generative capabilities but is notoriously difficult to train and is prone to instability and mode collapse. Flow-based models offer precise likelihood estimation and high-quality generation but require significant computational resources and involve complex model designs. Diffusion models excel in generating high-quality, detailed output, offering flexibility and solid probabilistic foundations, but suffer from long training times and high computational costs. However, many variants and improvements of diffusion models have mitigated these issues, as described in Sec.~\ref{DM-Variants}.
\vspace{-0.3cm}
\section{Overview and Categorization} \label{sec:overview and categorization}

\begin{table*}[]
    \caption{Summary and main papers of the diffusion models for time series and spatio-temporal data modeling. Red indicates \colorbox{red!20}{univariate time series}, blue-violet is \colorbox{blue!20}{multivariate time series}, and yellow shows \colorbox{yellow!20}{spatio-temporal data}.} \vspace{-8pt}
    \label{tab:paper_list}
    \centering
    \resizebox{0.95\textwidth}{!}{

\begin{tabular}{llllllll}
\hline \hline
\textbf{Method} & \textbf{Data} & \textbf{Model} & \textbf{Task} & \textbf{Application} & \textbf{Institute} & \textbf{Venue} & \textbf{Year} \\ \hline
\rowcolor{red!20}
WaveGrad~\cite{chen2020wavegrad} & Univariate & DDPM & Generation & Audio & Google & ICLR & 2021 \\
\rowcolor{red!20}
DiffWave~\cite{kong2020diffwave} & Univariate & DDPM & Generation & Audio & UCSD & ICLR & 2021 \\
\rowcolor{red!20}
D-Va~\cite{koa2023diffusion} & Univariate & DDPM & Forecasting & Finance & NUS & CIKM & 2023 \\
\rowcolor{red!20}
DiffLoad~\cite{wang2023diffload} & Univariate & DDPM & Forecasting & Electricity & HKU & ArXiv & 2023 \\
\rowcolor{red!20}
DiffuASR~\cite{liu2023diffusion} & Univariate & DDPM & Generation & Recommendation & XJTU & CIKM & 2023 \\
\rowcolor{red!20}
DiffRec~\cite{wang2023diffusion} & Univariate & DDPM & Generation & Recommendation & NUS & SIGIR & 2023 \\
\rowcolor{blue!20}
TimeGrad~\cite{rasul2021autoregressive} & Multivariate & DDPM & Forecasting & General & Zalando & ICML & 2021 \\
\rowcolor{blue!20}
CARD~\cite{han2022card} & Multivariate & DDPM & Classification & General & UT-Austin & NeurIPS & 2022 \\
\rowcolor{blue!20}
BVAE~\cite{li2022generative} & Multivariate & DDPM & Forecasting & General & Baidu & NeurIPS & 2023 \\
\rowcolor{blue!20}
SSSD~\cite{alcaraz2022diffusion} & Multivariate & DDPM & Imputation & General & Oldenburg & TMLR & 2023 \\
\rowcolor{blue!20}
DA-TASWDM~\cite{xu2023density} & Multivariate & DDPM & Imputation & Healthcare & HKBU & CIKM & 2023 \\
\rowcolor{blue!20}
DiffEEG~\cite{shu2023data} & Multivariate & DDPM & Forecasting & Healthcare & USTC & ArXiv & 2023 \\
\rowcolor{blue!20}
Tosato   et al.~\cite{tosato2023eeg} & Multivariate & DDPM & Classification & Healthcare & Tilburg & Synapsium & 2023 \\
\rowcolor{blue!20}
DiffCharge~\cite{li2024diffcharge} & Multivariate & DDPM & Generation & Electricity & HKUST & ArXiv & 2023 \\
\rowcolor{blue!20}
Yang   et al.~\cite{yang2023diffusion} & Multivariate & DDPM & Imputation\&Anomaly   Detection & AIOps & Microsoft & ESEC/FSE & 2023 \\
\rowcolor{blue!20}
AnoDDPM~\cite{sui2024anomaly} & Multivariate & DDPM & Anomaly   Detection & General & Beihang & IEEE   Sens. J. & 2024 \\
\rowcolor{blue!20}
DiffShape~\cite{liu2024diffusion2} & Multivariate & DDPM & Classification & General & SCUT & AAAI & 2024 \\
\rowcolor{yellow!20}
STPP~\cite{yuan2023spatio} & Spatio-temporal & DDPM & Forecasting & General   Event & THU & KDD & 2023 \\
\rowcolor{yellow!20}
ERDiff~\cite{wang2024extraction} & Spatio-temporal & DDPM & Alignment & General & GIT & NeurIPS & 2023 \\
\rowcolor{yellow!20}
SpecSTG~\cite{lin2024specstg} & Spatio-temporal & DDPM & Forecasting & Transportation & USYD & ArXiv & 2024 \\
\rowcolor{yellow!20}
DST-DDPM~\cite{chen2024quantifying} & Spatio-temporal & DDPM & Forecasting & Environment & HKUST & ENVIRON   RES & 2024 \\
\rowcolor{yellow!20}
UTD-PTP~\cite{10489838} & Spatio-temporal & DDPM & Forecasting & Transportation & BIT & IEEE   Sens. J. & 2024 \\ \hline
\rowcolor{red!20}
SGMSE~\cite{welker2022speech} & Univariate & Score-based & Generation & Audio & UHH & Interspeech & 2022 \\
\rowcolor{red!20}
DeScoD-ECG~\cite{li2023descod} & Univariate & Score-based & Denoising & Healthcare & U   of A & IEEE   J BIOMED HEALTH & 2023 \\
\rowcolor{blue!20}
ScoreGrad~\cite{yan2021scoregrad} & Multivariate & Score-based & Forecasting & General & BIT & ArXiv & 2021 \\
\rowcolor{blue!20}
Bilo et al.~\cite{bilovs2022modeling} & Multivariate & Score-based & Forecasting & General & TUM & ICML & 2023 \\
\rowcolor{blue!20}
StoRM~\cite{lemercier2023storm} & Multivariate & Score-based & Denoising\&Generation & Audio &  UHH & IEEE-ACM   T AUDIO SPE & 2023 \\
\rowcolor{blue!20}
Risk-sensitive   SDE~\cite{li2024risk} & Multivariate & Score-based & Generation & General & Cambridge & ArXiv & 2024 \\
\rowcolor{blue!20}
TimeADDM~\cite{hu2024unsupervised} & Multivariate & Score-based & Anomaly   Detection & General & HFUT & ICASSP & 2024 \\
\rowcolor{yellow!20}
Sasdim~\cite{zhang2023sasdim} & Spatio-temporal & Score-based & Imputation & General & CSU & ArXiv & 2023 \\
\rowcolor{yellow!20}
Dyffusion~\cite{ruhling2024dyffusion} & Spatio-temporal & Score-based & Forecasting & General & UCSD & NeurIPS & 2023 \\ \hline
\rowcolor{red!20}
Lu et al.~\cite{lu2022conditional} & Univariate & Conditional & Generation & Audio & CMU & ICASSP & 2022 \\
\rowcolor{red!20}
PulseDiff~\cite{jenkins2023improving} & Univariate & Conditional & Imputation & Healthcare & IC & ArXiv & 2023 \\
\rowcolor{red!20}
DR-DiffuSE~\cite{tai2023revisiting} & Univariate & Conditional & Generation & Audio & UESTC & AAAI & 2023 \\
\rowcolor{red!20}
Dose~\cite{tai2024dose} & Univariate & Conditional & Generation & Audio & UESTC & NeurIPS & 2023 \\
\rowcolor{red!20}
FUSE~\cite{yang2024pre} & Univariate & Conditional & Generation & Audio & Oxford & Interspeech & 2024 \\
\rowcolor{red!20}
DiffsFormer~\cite{gao2024diffsformer} & Univariate & Conditional & Imputation\&Generation & Finance & USTC & ArXiv & 2024 \\
\rowcolor{red!20}
DreamRec~\cite{yang2024generate} & Univariate & Conditional & Generation & Recommendation & USTC & NeurIPS & 2024 \\
\rowcolor{blue!20}
CSDI~\cite{tashiro2021csdi} & Multivariate & Conditional & Imputation & General & Stanford & NeurIPS & 2021 \\
\rowcolor{blue!20}
TimeDiff~\cite{shen2023non} & Multivariate & Conditional & Forecasting & General & HKUST & ICML & 2023 \\
\rowcolor{blue!20}
DiffAD~\cite{xiao2023imputation} & Multivariate & Conditional & Anomaly   Detection & General & HENU & KDD & 2023 \\
\rowcolor{blue!20}
D$^3$R~\cite{wang2024drift} & Multivariate & Conditional & Anomaly   Detection & General & BUPT & NeurIPS & 2023 \\
\rowcolor{blue!20}
MIDM~\cite{wang2023observed} & Multivariate & Conditional & Imputation & General & USTC & KDD & 2023 \\
\rowcolor{blue!20}
MEDiC~\cite{sharma2023medic} & Multivariate & Conditional & Imputation & Healthcare & IIT & NeurIPS & 2023 \\
\rowcolor{blue!20}
DiffPLF~\cite{li2024diffplf} & Multivariate & Conditional & Forecasting & Electricity & HKUST(GZ) & ArXiv & 2024 \\
\rowcolor{blue!20}
ImDiffusion~\cite{chen2023imdiffusion} & Multivariate & Conditional & Anomaly   Detection & General & PKU & VLDB & 2024 \\
\rowcolor{blue!20}
DiffDA~\cite{huang2024diffda} & Multivariate & Conditional & Forecasting & Climate & ETH   Zurich & ArXiv & 2024 \\
\rowcolor{blue!20}
DiffSTOCK~\cite{daiya2024diffstock} & Multivariate & Conditional & Forecasting & Finance & Purdue & ICASSP & 2024 \\
\rowcolor{blue!20}
RF-Diffusion~\cite{chi2024rf} & Multivariate & Conditional & Generation & Network & THU & MobiCom & 2024 \\
\rowcolor{yellow!20}
DiffSTG~\cite{wen2023diffstg} & Spatio-temporal & Conditional & Forecasting & General & BJTU & SIGSPATIAL & 2023 \\
\rowcolor{yellow!20}
PriSTI~\cite{liu2023pristi} & Spatio-temporal & Conditional & Imputation & General & Beihang & ICDE & 2023 \\
\rowcolor{yellow!20}
DiffUFlow~\cite{zheng2023diffuflow} & Spatio-temporal & Conditional & Forecasting & Transportation & CSU & CIKM & 2023 \\
\rowcolor{yellow!20}
DiffTraj~\cite{zhu2024difftraj} & Spatio-temporal & Conditional & Generation & Transportation & SUSTech & NeurIPS & 2023 \\
\rowcolor{yellow!20}
ControlTraj~\cite{zhu2024controltraj} & Spatio-temporal & Conditional & Generation & Transportation & HKUST(GZ) & arXiv & 2024 \\
\rowcolor{yellow!20}
Diff-RNTraj~\cite{wei2024diff} & Spatio-temporal & Conditional & Generation & Transportation & BJTU & ArXiv & 2024 \\ \hline
\rowcolor{red!20}
Evans   et al.~\cite{evans2024long} & Univariate & LDM & Generation & Audio & Stability   AI & ArXiv & 2024 \\
\rowcolor{blue!20}
LDT~\cite{feng2024latent} & Multivariate & LDM & Forecasting & General & NTU & AAAI & 2024 \\
\rowcolor{blue!20}
Aristimunha   et al.~\cite{aristimunha2023synthetic} & Multivariate & LDM & Generation & Healthcare & UPSaclay & NeurIPS & 2023 \\
\rowcolor{blue!20}
TSDM~\cite{pei2023improved} & Multivariate & DDIM & Imputation\&Anomaly   Detection & Electricity & HUST & ArXiv & 2023 \\
\rowcolor{yellow!20}
LADM~\cite{lv2024learning} & Spatio-temporal & LDM & Forecasting & Transportation & XJU & IEEE   T INSTRUM MEAS & 2024 \\ \hline  \hline
\end{tabular}} \vspace{-13pt}
\end{table*}

\tikzstyle{my-box}=[
    rectangle,
    draw=hidden-draw,
    rounded corners,
    text opacity=1,
    minimum height=1.5em,
    minimum width=5em,
    inner sep=2pt,
    align=center,
    fill opacity=.5,
    line width=0.8pt,
]
\tikzstyle{leaf}=[my-box, minimum height=1.5em,
    fill=hidden-pink!80, text=black, align=left, font=\normalsize,
    inner xsep=2pt,
    inner ysep=4pt,
    line width=0.8pt,
]
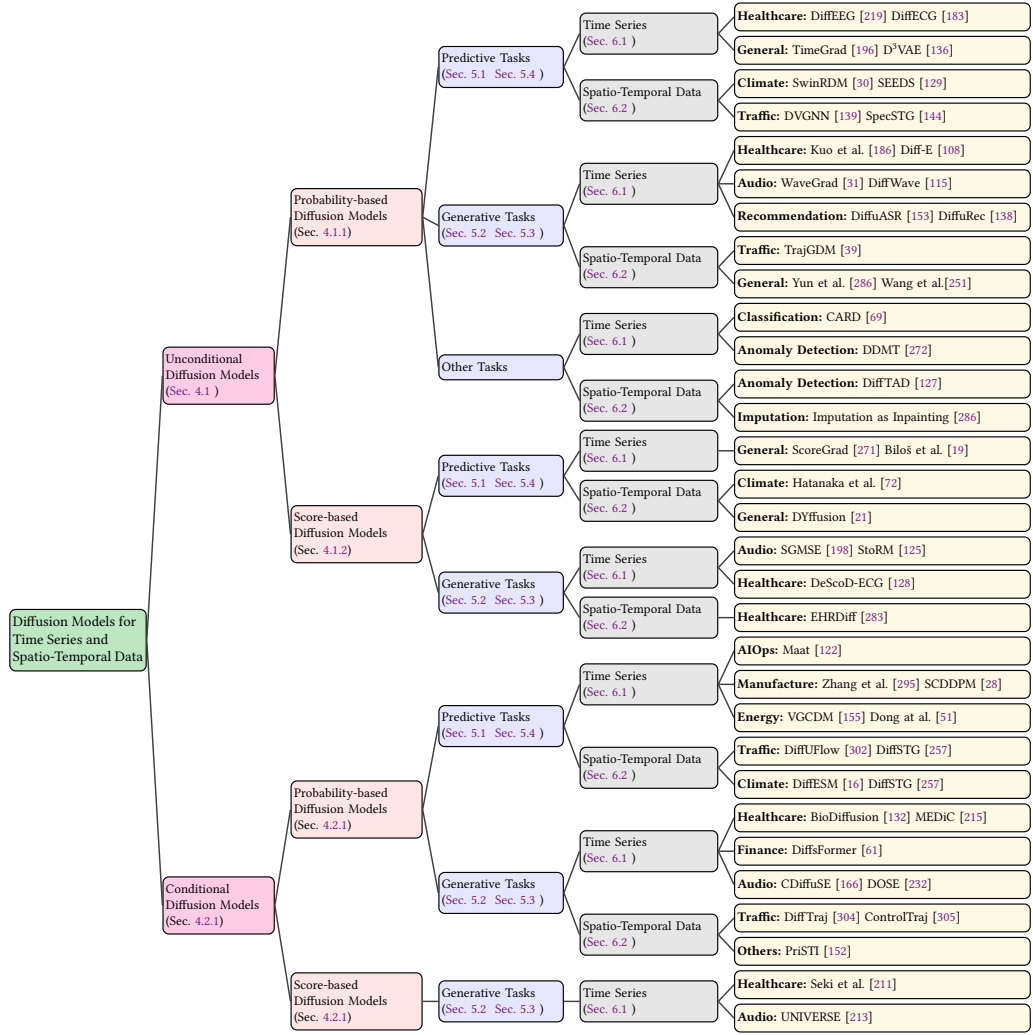
\begin{figure*}[!t]
    \centering
    \resizebox{0.9\textwidth}{!}{
        \begin{forest}
            edge={thick}
            forked edges,
            for tree={
                fill=level0!80,
                grow=east,
                reversed=true,
                anchor=base west,
                parent anchor=east,
                child anchor=west,
                base=left,
                font=\large,
                rectangle,
                draw=hidden-draw,
                rounded corners,
                align=left,
                minimum width=4em,
                edge+={darkgray, line width=1pt},
                s sep=3pt,
                inner xsep=2pt,
                inner ysep=3pt,
                line width=0.8pt,
                ver/.style={rotate=90, child anchor=north, parent anchor=south, anchor=center},
            },
            where level=1{text width=8em,font=\normalsize,fill=pink!20,}{},
            where level=2{text width=9.5em,font=\normalsize,fill=red!10,}{},
            where level=3{text width=9em,font=\normalsize,fill=blue!10,}{},
            where level=4{text width=10em,font=\normalsize,fill=gray!20,}{},
            where level=5{text width=5em,font=\normalsize,}{},
            [
                Diffusion Models for \\ Time Series and \\ Spatio-Temporal Data
                [
                    Unconditional \\ Diffusion Models  \\
                    (\shortautoref{subsec:normal dm})
                    [
                       Probability-based \\Diffusion Models \\
                       (Sec. \ref{subsec:Probability-Based Model})
                        [
                            Predictive Tasks \\(\shortautoref{subsec:forecasting}\ \shortautoref{subsec:anomaly-detection})
                                [
                                Time Series \\(\shortautoref{subsec:time-series-data})
                                    [ 
                                    \textbf{Healthcare:} DiffEEG~\cite{shu2023data} DiffECG~\cite{neifar2023diffecg},leaf, text width=22em
                                    ]
                                    [ 
                                        \textbf{General:} TimeGrad~\cite{rasul2021autoregressive} D$^3$VAE~\cite{li2022generative},leaf, text width=22em
                                    ]
                                ]
                                [
                                Spatio-Temporal Data \\(\shortautoref{subsec:spatio-temporal-data})
                                    [ 
                                        \textbf{Climate:} SwinRDM~\cite{chen2023swinrdm} SEEDS~\cite{li2023seeds},leaf, text width=22em
                                    ]
                                    [ 
                                        \textbf{Traffic:} DVGNN~\cite{liang2023dynamic} SpecSTG~\cite{lin2024specstg},leaf, text width=22em
                                    ]
                                ]
                        ]
                        [
                            Generative Tasks \\(\shortautoref{subsec:generation}\ \shortautoref{subsec:imputation})
                            [
                                Time Series \\(\shortautoref{subsec:time-series-data})
                                [ 
                                    \textbf{Healthcare:} Kuo et al.~\cite{nicholas2023synthetic} Diff-E~\cite{kim2023diff},leaf, text width=22em
                                ]
                                [ 
                                    \textbf{Audio:} WaveGrad~\cite{chen2020wavegrad} DiffWave~\cite{kong2020diffwave},leaf, text width=22em
                                ]
                                [ 
                                    \textbf{Recommendation:} DiffuASR~\cite{liu2023diffusion} DiffuRec~\cite{li2023diffurec},leaf, text width=22em
                                ]
                            ]
                            [
                                Spatio-Temporal Data \\(\shortautoref{subsec:spatio-temporal-data})
                                [ 
                                    \textbf{Traffic:} TrajGDM~\cite{chu2024simulating},leaf, text width=22em
                                ]
                                [ 
                                    \textbf{General:} Yun et al.~\cite{yun2023imputation} Wang et al.\cite{wang2024extraction},leaf, text width=22em
                                ]
                            ]
                        ]
                        [
                            Other Tasks
                            [
                                Time Series \\(\shortautoref{subsec:time-series-data})
                                [ 
                                    \textbf{Classification:} CARD~\cite{han2022card},leaf, text width=22em
                                ]
                                [ 
                                    \textbf{Anomaly Detection:} DDMT~\cite{yang2023ddmt},leaf, text width=22em
                                ]
                            ]
                            [
                                Spatio-Temporal Data \\(\shortautoref{subsec:spatio-temporal-data})
                                [ 
                                    \textbf{Anomaly Detection:} DiffTAD~\cite{LI2024111387},leaf, text width=22em
                                ]
                                [ 
                                    \textbf{Imputation:} Imputation as Inpainting~\cite{yun2023imputation},leaf, text width=22em
                                ]
                            ]
                        ]
                    ]
                    [
                        Score-based \\ Diffusion Models \\
                        (Sec. \ref{subsec:Score-Based Model})
                        [
                            Predictive Tasks \\(\shortautoref{subsec:forecasting}\ \shortautoref{subsec:anomaly-detection}) 
                            [
                                Time Series \\(\shortautoref{subsec:time-series-data})
                                [ 
                                    \textbf{General:} ScoreGrad~\cite{yan2021scoregrad} Biloš et al.~\cite{bilovs2022modeling},leaf, text width=22em
                                ]
                            ]
                            [
                                Spatio-Temporal Data \\(\shortautoref{subsec:spatio-temporal-data})
                                [ 
                                    \textbf{Climate:} Hatanaka et al.~\cite{hatanaka2023diffusion},leaf, text width=22em
                                ]
                                [ 
                                    \textbf{General:} DYffusion~\cite{cachay2023dyffusion},leaf, text width=22em
                                ]
                            ]
                        ]
                        [
                            Generative Tasks \\(\shortautoref{subsec:generation}\ \shortautoref{subsec:imputation})
                            [
                                Time Series \\(\shortautoref{subsec:time-series-data})
                                [ 
                                    \textbf{Audio:} SGMSE~\cite{richter2023speech} StoRM~\cite{lemercier2023storm},leaf, text width=22em
                                ]
                                [ 
                                    \textbf{Healthcare:} DeScoD-ECG~\cite{li2023descod},leaf, text width=22em
                                ]
                            ]
                            [
                                Spatio-Temporal Data \\(\shortautoref{subsec:spatio-temporal-data})
                                [ 
                                    \textbf{Healthcare:} EHRDiff~\cite{yuan2023ehrdiff},leaf, text width=22em
                                ]
                            ]
                        ]
                    ]
                ]
                [                
                    Conditional \\ Diffusion Models \\
                    (Sec. \ref{sub:conditional dm})
                    [
                        Probability-based \\Diffusion Models \\
                        (Sec. \ref{sub:conditional dm})
                        [
                            Predictive Tasks \\(\shortautoref{subsec:forecasting}\ \shortautoref{subsec:anomaly-detection})
                            [
                                Time Series \\(\shortautoref{subsec:time-series-data})
                                [
                                    \textbf{AIOps:} Maat~\cite{lee2023maat},leaf, text width=22em
                                ]
                                [
                                    \textbf{Manufacture:} Zhang et al.~\cite{zhang2023multi} SCDDPM~\cite{chen2024online},leaf, text width=22em
                                ]
                                [
                                    \textbf{Energy:} VGCDM~\cite{liu2023generating} Dong at al.~\cite{dong2023short},leaf, text width=22em
                                ]
                            ]
                            [
                                Spatio-Temporal Data \\(\shortautoref{subsec:spatio-temporal-data})
                                [ 
                                    \textbf{Traffic:} DiffUFlow~\cite{zheng2023diffuflow} DiffSTG~\cite{wen2023diffstg},leaf, text width=22em
                                ]
                                [ 
                                    \textbf{Climate:} DiffESM~\cite{bassetti2023diffesm} DiffSTG~\cite{wen2023diffstg},leaf, text width=22em
                                ]
                            ]
                        ]
                        [
                            Generative Tasks \\(\shortautoref{subsec:generation}\ \shortautoref{subsec:imputation})
                            [
                                Time Series \\(\shortautoref{subsec:time-series-data})
                                [ 
                                    \textbf{Healthcare:} BioDiffusion~\cite{li2024biodiffusion} MEDiC~\cite{sharma2023medic}, leaf, text width=22em
                                ]
                                [ 
                                    \textbf{Finance:} DiffsFormer~\cite{gao2024diffsformer} ,leaf, text width=22em
                                ]
                                [ 
                                    \textbf{Audio:} CDiffuSE~\cite{lu2022conditional} DOSE~\cite{tai2024dose},leaf, text width=22em
                                ]
                            ]
                            [
                                Spatio-Temporal Data \\(\shortautoref{subsec:spatio-temporal-data})
                                [ 
                                    \textbf{Traffic:}  DiffTraj~\cite{zhu2024difftraj} ControlTraj~\cite{zhu2024controltraj},leaf, text width=22em
                                ]
                                [ 
                                    \textbf{Others:}  PriSTI~\cite{liu2023pristi},leaf, text width=22em
                                ]
                            ]
                        ]
                    ]
                    [
                        Score-based \\ Diffusion Models \\
                        (Sec. \ref{sub:conditional dm})
                        [
                            Generative Tasks \\(\shortautoref{subsec:generation}\ \shortautoref{subsec:imputation})
                            [
                                Time Series \\(\shortautoref{subsec:time-series-data})
                                [ 
                                    \textbf{Healthcare:} Seki et al.~\cite{seki2023imputing},leaf, text width=22em
                                ]
                                [ 
                                    \textbf{Audio:} UNIVERSE~\cite{serra2022universal},leaf, text width=22em
                                ]
                            ]
                      ]
                    ]
                ]
            ]
        \end{forest}
    }
\caption{A comprehensive taxonomy of diffusion models for time series and spatio-temporal data, categorized according to methodologies (i.e., unconditional vs. conditional), tasks (e.g., predictive versus generative), data types, and applications.}
\label{fig:taxonomy} \vspace{-0.4cm}
\end{figure*}

This section presents an overview and classification of diffusion models for addressing challenges in time series and spatio-temporal data analysis. Our survey organizes the discussion along four primary dimensions: categories of diffusion models, types of tasks, data modalities, and practical applications\footnote{All summarized and categorized papers document: \url{https://docs.google.com/spreadsheets/d/1X5ujA-yXCzkzz4Uu9ErP8mu_Jg2zWk4fwZrwrMcqMoI}.}. A comprehensive summary of notable related works is depicted in \shortautoref{fig:taxonomy}. We categorize the existing literature into two primary groups: \emph{\textbf{unconditioned}} and \emph{\textbf{conditioned}} diffusion models, focusing on time series and spatio-temporal data.

In the unconditioned category, diffusion models operate in an unsupervised manner to generate data samples without the need for supervision signals. This setting represents the foundational approach for analyzing time series and spatio-temporal data. Within this category, literature can be further divided into \emph{\textbf{probability-based}} and \emph{\textbf{score-based}} diffusion models. Examples include denoised diffusion probabilistic models (DDPMs)~\cite{ho2020denoising} and score-based stochastic differential equations (Score SDEs)~\cite{song2020score,song2019generative}, as introduced in \shortautoref{sec:background}. Research in this category is broadly organized into two task groups: \emph{\textbf{predictive}} and \emph{\textbf{generative}} tasks. Predictive tasks typically involve forecasting and anomaly detection, leveraging historical data and patterns to anticipate current and/or future events. Generative tasks, conversely, focus on identifying patterns within extensive datasets to generate new content, such as time series imputation and augmentation. Methods are developed for both primary data modalities: \emph{\textbf{time series}} and \emph{\textbf{spatio-temporal data}}, catering to a wide range of applications across various sectors, including healthcare, energy, climate, traffic, and more.

In the conditioned category, diffusion models are tailored for conditioned analysis of time series and spatio-temporal data. Empirical studies have shown that conditional generative models, which utilize data labels, are easier to train and yield superior performance compared to their unconditional counterparts~\cite{bao2022conditional}. In this context, labels (a.k.a. conditions) often derive from various sources, such as extracted short-term trends~\cite{shen2023non} and urban flow maps~\cite{zheng2023diffuflow}, to enhance model inferences. This category embraces both probability-based and score-based diffusion models for predictive and generative tasks, offering a fresher perspective on leveraging diffusion models to tackle practical challenges in time series and spatio-temporal data analysis under specific constraints.

Building on the foundational understanding of model categories, task types, data modalities, and application domains, we delve deeper into the exploration of diffusion models for time series and spatio-temporal data analysis across the following sections. Each section is designed to unpack the complexities and nuances inherent in the application of diffusion models, providing a comprehensive overview from multiple perspectives. In \shortautoref{sec: Models}, we explore diffusion model landscapes, highlighting distinctions between unconditioned and conditioned approaches and their implications. \shortautoref{sec: Task} analyzes tasks from the predictive and generative viewpoints, detailing specific functions such as forecasting, generation, anomaly detection, and data imputation. \shortautoref{sec: Modality} examines the data modalities, differentiating between time series and spatio-temporal data to outline model challenges and applicability. Lastly, \shortautoref{sec:applications} extends the discussion to application fields, demonstrating diffusion models' utility across sectors like healthcare, traffic, sequential recommendation, climate, energy, and audio. This structured exploration aims to equip readers with an in-depth understanding of the potential and current state of diffusion models for addressing complex time series and spatio-temporal data challenges.
\vspace{-0.2cm}
\section{Model Perspective} \label{sec: Models} 

In this section, we will analyze how to use diffusion models for time series and spatio-temporal data from a model perspective. Specifically, we will focus on standard diffusion models (DDPM and score SDE) and improved diffusion models (conditional diffusion model, LDM, DDIM, and others). In addition to the above content, we will also discuss in detail an important issue in the model section, namely, the denoisers in the diffusion model.

\vspace{-0.2cm}
\subsection{Standard Diffusion Model}\label{subsec:normal dm}
Standard diffusion models include probability-based model, i.e., DDPM and score-based models, score SDE, which have been described in detail in Sec.~\ref{subsubsec:DDPM} and Sec.~\ref{subsubsec:SDE}, respectively. Currently, diffusion models based on these two are also the most common methods for time series and spatio-temporal data analysis.

\subsubsection{Probability-Based Model}\label{subsec:Probability-Based Model}
Based on the probabilistic-based standard model, we mainly introduce DDPM there, which uses a \emph{\textbf{discrete}} framework, decomposing the diffusion process into a fixed number of steps. This method adds noise to the data at discrete intervals and then learns to reverse this process, generating data from noise. Both the training and inference process of the model are conducted in these discrete steps, making the process essentially stepwise and iterative.

TimeGrad is among the most classic DDPM-based time series forecasting methods, identified as a score-matching model that, through rigorous validation, has been proven effective in real-world datasets comprising thousands of interconnected dimensions~\cite{rasul2021autoregressive}. D3VAE proposes a bidirectional variational auto-encoder that integrates diffusion, denoising, and disentanglement to enhance time series data without introducing extraneous uncertainties. This model marries multi-scale denoising score matching with bi-directional variational autoencoding for forecasting tasks~\cite{li2022generative}. In a similar vein, TSDiff leverages implicit probability densities to iteratively refine the base forecasts, thereby addressing three distinct tasks: forecasting, refining, and generating synthetic data~\cite{kollovieh2024predict}. For the universal challenge of anomaly detection, Pintilie et al. have proposed two multivariate time series anomaly detection algorithms based on diffusion models, securing leading outcomes~\cite{pintilie2023time}. Concurrently, D$^3$R tackles temporal anomaly detection drifts through decomposition and reconstruction, employing data-time mix-attention for dynamic decomposition alongside diffusion models for end-to-end training, thus overcoming the limitations imposed by local sliding windows and unstable data scenarios~\cite{wang2024drift}. CARD presents a denoising diffusion-based generative model alongside a pretrained conditional mean estimator to serve both classification and regression tasks~\cite{han2022card}. TS-Diffusion is specifically designed for complex sequences marked by irregular sampling, missingness, and extensive temporal feature dimensions, introducing a comprehensive model consisting of an ODE encoder, a representational learning module, and an ODE decoder to handle such intricate time series~\cite{li2023ts}. Furthermore, WaveGrad~\cite{chen2020wavegrad}, DiffWave~\cite{kong2020diffwave}, and DiffuASR~\cite{liu2023diffusion} have each successfully applied DDPM to waveforms, audio generation, and sequence recommendation, respectively.

On the other hand, in the field of spatio-temporal data, significant advances have been made with DDPM predominantly starting from 2023, indicating an exciting opportunity for more applications and theoretical breakthroughs. Yun et al. have put forward "Imputation as Inpainting," which integrates an unconditional diffusion model based on graph neural networks to first forecast complete spatio-temporal data. This approach adjusts the generation process by sampling in unobserved areas using the information from observed data, thus estimating missing values in spatio-temporal data~\cite{yun2023imputation}. For traffic and specifically trajectory forecasting, the adoption of graph-based DDPM methods is becoming more common. For example, MID++ leverages a GNN-based ALEncoder to extract spatio-temporal features and improves the training process with an important sampling strategy, leading to better trajectory forecasting~\cite{yao2023graph}. In a similar vein, DiffTAD incorporates a Transformer-based decoupled time and space encoder to model spatial interactions among vehicles and conducts anomaly detection by evaluating the differences between query trajectories and their reconstructions~\cite{li2024difftad}. Moreover, SpecSTG tackles the challenges of insufficient representation of spatial network features and the inability to detect unexpected fluctuations in future observations in traffic flow forecasting. It achieves this by converting the learning process to the spectral domain, generating Fourier representations of future time series that carry spatial information~\cite{lin2024specstg}.

\subsubsection{Score-Based Model}\label{subsec:Score-Based Model}
Unlike the discrete DDPM, score-based SDE models represent diffusion and the reverse process in a \emph{\textbf{continuous}} form using stochastic differential equations, thereby covering continuous time. This approach allows for a theoretically more flexible and in-depth treatment of the diffusion process, capable of generating samples at any point in time, rather than being confined to fixed steps.

ScoreGrad and TimeGrad share similar goals and can be described as SDE-based TimeGrad models. They expand the diffusion process into a continuous spectrum and employ time series feature extraction modules along with a condition-based stochastic differential equation for score matching to facilitate forecasting~\cite{yan2021scoregrad}. This structure finds parallels in the work of Biloš et al.~\cite{bilovs2022modeling}. Additionally, recent studies by Crabbé et al. delve into representing time series data in the frequency domain using SDEs, showcasing how denoising score matching methods can facilitate diffusion modeling within the frequency domain based on differing time-frequency spaces~\cite{crabbe2024time}. TimeADDM introduces a score-based diffusion model for unsupervised anomaly detection in multivariate time series, applying diffusion steps to representations that encapsulate global time correlations through recurrent embeddings and designing a suite of reconstruction strategies to compute anomaly scores at various diffusion intensities~\cite{hu2024unsupervised}. Moreover, Li et al. extensively discuss the impact of noise samples in SDEs on data quality and the corresponding robustness of the models~\cite{li2024risk}.

From an application standpoint, score-based diffusion models have shown great promise in continuous data fields such as healthcare and acoustic data. EHRDiff~\cite{yuan2023ehrdiff} and DeScoD-ECG~\cite{li2023descod} leverage score-based diffusion models for realistic EHR synthesis and ECG data generation, respectively, incorporating algorithms or structures pertinent to the features of medical data, thus improving applicability in real-world settings. In audio-related applications, the focus has been on speech enhancement and dereverberation, with initial proposals by Welker et al.~\cite{welker2022speech} and Richter et al.~\cite{richter2023speech} for employing score-based diffusion models for speech enhancement, transforming time domain speech data into time-frequency information, and treating them as image input for models. StoRM introduces a stochastic resampling method, utilizing forecasting from a predictive model as guidance for subsequent diffusion steps, achieving higher-quality samples under complex conditions with high signal-to-noise ratios using fewer diffusion steps and more streamlined models~\cite{lemercier2023storm}. Similarly, Lay et al. emphasize improving model efficiency and optimizing model parameters, proposing a Brownian bridge-based forward process to reconcile the gap between the forward process's end distribution and the prior distribution used in inference for the reverse process~\cite{lay2023reducing}.

In the realm of spatio-temporal data, methods remain scarce, with Sasdim and DYffusion standing out. Sasdim, an adaptive noise scaling diffusion model, effectively performs spatio-temporal data imputation by capturing dynamic spatio-temporal dependencies through novel loss functions and global spatio-temporal convolution modules~\cite{zhang2023sasdim}. DYffusion, a dynamics-informed diffusion model for spatio-temporal data forecasting, distinguishes itself from probabilistic models by integrating temporal dynamics directly into the diffusion steps and training a stochastic, time-conditioned interpolator alongside a predictor network to simulate the forward and reverse processes of the standard diffusion model, thus striking a balance between performance and efficiency~\cite{ruhling2024dyffusion}.

\vspace{-0.2cm}
\subsection{Improved/Advanced Diffusion Model}\label{subsec:Imporved dm}
We have already discussed the improvement methods of the diffusion model in Sec.~\ref{CDM} and Sec.~\ref{DM-Variants}. In the following, we will analyze the methods of the more frequently used ones.

\subsubsection{Conditional Diffusion Model} \label{sub:conditional dm}
Compared to standard approaches, conditional diffusion models are able to use given \emph{\textbf{conditional information}} (e.g., different representations, different modalities, etc.) to steer the generative process and produce high-quality outputs that are tightly correlated with conditions~\cite{shen2023non}. More condition-based models are used for high-quality generation tasks.

CSDI and MIDM use conditional diffusion models for time series imputation~\cite{tashiro2021csdi, wang2023observed}. They utilize score-based and probability-based diffusion models conditioned on observed values and through explicit imputation training. They can leverage the correlations among observed values to further enhance the performance. Similar to CSDI, DiffAD and ImDiffusion employ a similar method based on conditional weight-incremental diffusion to enhance the imputation performance of missing values and are applied for time series anomaly detection~\cite{xiao2023imputation, chen2023imdiffusion}. These approaches preserve the information on the observed values and significantly improve the generation quality for stable anomaly detection. On the other hand, there are models that utilize different guiding information~\cite{fu2024creating}. For example, \cite{zuo2023unsupervised} adjusts the diffusion model based on statistical information such as mean, standard deviation, Z-scores, and skewness, thereby synthesizing sensor data. \cite{tai2024dose} proposes two different conditional enhancement techniques to enable the model to adaptively consider conditional information a priori, thus performing the speech enhancement task. MEDiC~\cite{sharma2023medic} introduces a class-conditional DDPM approach to generate synthetic EEG embeddings. VGCDM~\cite{liu2023generating} employs a pulse voltage-guided diffusion model along with a cross-attention mechanism for more efficient generation of electrical signals. Meanwhile, Wang et al. leverage multimodal information (images and text) as generative conditions to enhance the quality of real-time sales forecasts~\cite{wang2024multi}. DiffShape~\cite{liu2024diffusion2} introduces a self-supervised diffusion learning mechanism that uses real sub-sequences as conditions. By leveraging a large amount of unlabeled data, it enhances the similarity between learned shapelets and actual sub-sequences for classification.

For spatio-temporal data, most models adopt graph structures for representation, but the method of using conditions is similar to that of time series data~\cite{ling2024srndiff,liu2024intention}. PriSTI draws inspiration from CSDI of conditioning on the observed value, employing a conditional feature extraction module based on linear interpolation for generating missing data~\cite{liu2023pristi}. DiffTraj estimates noise levels accurately using various external factors, such as the region of the trip and departure time~\cite{zhu2024difftraj}. ControlTraj further extends DiffTraj with the constraint of road network structures~\cite{zhu2024controltraj}. Additionally, many works guide their models using graph structures or feature graphs as conditions. For instance, DiffSTG~\cite{wen2023diffstg} and USTD~\cite{hu2023towards} use historical graph signals and graph structures as conditions, combining the spatio-temporal learning capabilities of GNNs and the uncertainty measurement of diffusion models to enhance the performance of forecasting tasks. DiffUFlow~\cite{zheng2023diffuflow} utilizes extracted spatio-temporal feature graphs overlaid on coarse-grained flow graphs as a condition to guide the reverse process.

\subsubsection{Latent Diffusion Model}
The latent diffusion model (LDM) performs the diffusion process in a lower-dimensional latent space, which allows more \emph{\textbf{efficient}} generation or other tasks. It allows the model to handle more complex data distributions, while reducing computational resource consumption and maintaining output quality.

LDCast utilizes LDM for near-term precipitation forecasting, enhancing training stability, optimizing computational demands, and accurately representing the uncertainty of forecasting compared to GAN~\cite{leinonen2023latent}. CLDM presents a conditional latent diffusion model based on latent space mapping, which breaks down the generative task into deterministic forecasting and the generation of predictive error scenarios, thus increasing efficiency within the latent space~\cite{dong2023short}. Similarly, Aristimunha et al. suggest the use of LDM for generating sleep EEG signals, where the latent feature maps outputted by the encoders serve as input to the diffusion model, with the generated results obtained through the corresponding decoder~\cite{aristimunha2023synthetic}. In a similar manner, LADM proposes a method for spatio-temporal trajectory forecasting, where VAE acts as a generator and DDPM as a refiner~\cite{lv2024learning}. Furthermore, Feng et al. develop a latent diffusion transformer for time series forecasting, aimed at compressing multivariate time stamp patterns into succinct latent representations and efficiently generating authentic multivariate time stamp values in a continuous latent space~\cite{feng2024latent}.

\subsubsection{Other Variants of Diffusion Models}
There are also other methods that incorporate the different models above or use other more advanced diffusion models. For example, to accelerate generation, MedDiff for the first time utilizes DDIM with a new sampling strategy to generate high-dimensional large-scale electronic medical records~\cite{he2023meddiff}. Specifically, it incorporates a classifier to guide the sampling process, assigning high probabilities to the data with the correct labels. DCM is a diffusion-based causal inference model accelerated by DDIM, that more accurately captures counterfactual distributions under unmeasured confounding factors~\cite{shimizu2023diffusion}. Asperti et al. propose GED, which uses conditional DDIM integrated with additional information and a post-processing network for high-performance weather forecasting~\cite{asperti2023precipitation}. 

Besides, some algorithms have also been proposed that use a two-stage or a diffusion model to fuse with other models. TSDM adopts an improved two-stage diffusion model for identifying and reconstructing measurements with various uncertainties for power system measurement recovery, where the first stage includes a classifier-guided conditional anomaly detection component, and the second stage involves a diffusion-based measurement imputation component~\cite{pei2023improved}. Mueller proposes using attention-enhanced condition-guided DDIM to tackle sample imbalance and data scarcity issues, along with simple DDPM for signal denoising, thereby synthesizing effective machine fault data for efficient anomaly detection~\cite{mueller2024attention}. Similarly, Wong et al. utilize DDIM for load domain adaptation, further enhanced with CNN for bearing fault anomaly detection~\cite{wong2024denoising}.

\vspace{-0.2cm}
\subsection{Denoisers in the Diffusion Model}
Denoisers in the diffusion model serve as the core modules that transform noise into structured signals, with their architectures, inductive biases, and prior-guided designs fundamentally shaping the model’s generative capacity, temporal reasoning, and applicability to diverse inverse problems. In the following, we will provide a detailed analysis of these aspects respectively.

\subsubsection{Architecture of the Denoiser}
The denoiser serves as the backbone of diffusion models, determining their expressive power and cross-modal generalization. U-Net remains the architecture most widely used in diffusion models~\cite{ronneberger2015u,ho2020denoising,rombach2022high}, where its encoder–decoder design with skip connections supports multi-scale representation learning and effective propagation of global and local information. Models such as DiffuASR~\cite{liu2023diffusion} and DiffAD~\cite{xiao2023imputation} adapt U-Net with 1D convolutions or temporal dilations to better model sequential dependencies. Some works explore attention-based backbones for long-range dependency modeling, including CSDI~\cite{tashiro2021csdi} and PriSTI~\cite{liu2023pristi}, which enhance global context capture and scalability~\cite{zheng2023diffuflow}. Hybrid architectures further extend denoising to structured data. D$^3$R combines attention and diffusion for anomaly detection~\cite{wang2024drift}, while SpecSTG uses graph spectral convolutions for traffic forecasting~\cite{lin2024specstg}. These domain-aware designs show that integrating prior knowledge improves denoising efficiency and interpretability~\cite{wen2023diffstg}. Recent trends favor modular, cross-modal and lightweight denoisers, including foundation backbones with temporal or structural adaptations~\cite{ma2025efficient}.

\subsubsection{Inductive Bias in Temporal Denoisers}
The effectiveness of diffusion models relies strongly on the inductive biases in their denoiser architectures. U-Net structures leverage translation equivariance and hierarchical feature extraction for robust spatial representations~\cite{kadkhodaie2024generalization,kamb2025an}. For time series, analogous inductive priors must capture temporal continuity, periodicity, and causal dynamics. Hierarchical temporal architectures (e.g., Temporal U-Net, WaveGrad~\cite{chen2020wavegrad}) address multi-scale temporal patterns through progressive downsampling and upsampling. In spatio-temporal settings, CNN-based denoisers (e.g., DiffTraj~\cite{zhu2024difftraj}) learn long-term dependencies efficiently with skip connections, while graph-based denoisers (e.g., DiffSTG~\cite{wen2023diffstg}, DiffUFlow~\cite{zheng2023diffuflow}) incorporate structural priors via graph convolutions. Emerging temporal transformers (e.g., DiffUFlow~\cite{zheng2023diffuflow}, PriSTI~\cite{liu2023pristi}) unify causal and hierarchical biases through self-attention with relative positional encodings.

\subsubsection{Denoiser-as-Prior in Temporal Inverse Problems}

Beyond the full diffusion sampling framework, many inverse problems can be solved by directly leveraging the denoiser (or score function) as a learned prior. This idea has been widely explored under the Plug-and-Play and Score Prior paradigms, where the denoiser acts as a proximal operator for iterative reconstruction~\cite{kadkhodaie2020solving,milanfar2025denoising}. For the time series field, a similar trend has emerged. Models such as ImDiffusion~\cite{chen2023imdiffusion} and SADI~\cite{islam2025self} employ the denoiser alone to restore missing segments, achieving high-quality reconstructions without full stochastic sampling. This approach reduces computational overhead and enhances interpretability by decoupling temporal prior learning from generative sampling. The denoiser-as-prior framework therefore provides a promising bridge between generative diffusion modeling and classical temporal signal restoration.
\vspace{-0.2cm}
\section{Task Perspective} \label{sec: Task} 

In this section, we will explore the use of diffusion models in different tasks, such as forecasting, generation, imputation, and anomaly detection, also highlighting their effectiveness in complex time series and spatio-temporal data analysis across various domains.

\vspace{-0.2cm}
\subsection{Forecasting}\label{subsec:forecasting}

The field of time series forecasting has seen significant advancements with the incorporation of diffusion models. TimeGrad \cite{rasul2021autoregressive} and D3VAE \cite{li2022generative} both employ diffusion probabilistic models to enhance forecasting, with TimeGrad focusing on autoregressive techniques (i.e., RNN) for probabilistic forecasting and D3VAE introducing a bidirectional variational auto-encoder that includes diffusion and denoising processes. This generative approach is further extended in the work by D-Va \cite{koa2023diffusion}, which addresses the stochastic nature of stock price data through a deep hierarchical VAE combined with diffusion probabilistic techniques. Meanwhile, the study presented in \cite{bilovs2022modeling} takes a different approach by modeling temporal data as continuous functions, allowing the handling of irregularly sampled data. Building upon the diffusion model, TimeDiff \cite{shen2023non} introduces novel conditioning mechanisms, future mixup, and autoregressive initialization, to improve time series forecasting. Lastly, the research in \cite{kollovieh2024predict} explores task-agnostic unconditional diffusion models, proposing TSDiff, which employs a self-guidance mechanism for versatile time series applications. 

For spatio-temporal data, various diffusion model-based approaches have been proposed to tackle complex forecasting problems. DiffSTG \cite{wen2023diffstg} is the first work that generalizes denoising diffusion probabilistic models to spatio-temporal graphs, aiming to model complex spatio-temporal dependencies and intrinsic uncertainties within spatio-temporal graph data for better forecasting.  DYffusion \cite{ruhling2024dyffusion} introduces a framework that trains a stochastic, time-conditioned interpolator and a forecaster network to perform multi-step and long-range probabilistic forecasting for spatio-temporal data. On the other hand, DiffUFlow \cite{zheng2023diffuflow} focuses on urban data, which aims to address the challenge of fine-grained flow inference. DSTPP \cite{yuan2023spatio} provides a novel parameterization for spatio-temporal point processes. Meanwhile, SwinRDM \cite{chen2023swinrdm} and DOT \cite{lin2023origin} demonstrate the adaptability of diffusion models in improving the quality of weather forecasts and travel time estimations, showcasing the wide applicability of these models across different domains and tasks. These works highlight the growing impact and potential of diffusion models in advancing spatio-temporal data analysis.

\vspace{-0.2cm}
\subsection{Generation}\label{subsec:generation}
Inspired by the powerful ability of the diffusion model in high-dimensional distribution learning, an intuitive usage is applying the learned diffusion model for data generation. Currently, a variety of diffusion models have been proposed for generating audio data. WaveGrad \cite{chen2020wavegrad} is a pioneering work that utilizes gradient-based sampling to generate high-fidelity audio waveforms, marking a significant advancement in audio synthesis. Similarly, DiffWave \cite{kong2020diffwave} employs a non-autoregressive approach to achieve efficient and high-quality raw audio synthesis through a Markov chain process. Both WaveGrad and DiffWave are part of the same research lineage, leveraging the power of diffusion models to create complex waveforms from simple noise distributions. DOSE \cite{tai2024dose} takes a different approach by focusing on the speech enhancement task, which introduces a model-agnostic method that integrates condition information into the diffusion process. Besides, diffusion model is also used in sequential recommendation. DiffuASR \cite{liu2023diffusion} proposes a diffusion-based sequence generation framework to address data sparsity and long-tail user problems in sequential recommendation systems. It introduces a sequential U-Net designed for discrete sequence generation tasks and utilizes two guide strategies to assimilate preferences between generated and original sequences. DreamRec \cite{yang2024generate} employs a Transformer encoder to create guidance representations as conditions in the diffusion process. 

In the realm of spatio-temporal data, diffusion models are adopted to generate trajectory data. For instance, DiffTraj \cite{zhu2024difftraj,zhu2024controltraj} proposes the first effort that generates high-quality human trajectories with an unconditioned model, with the motivation to protect privacy. Meanwhile, \cite{lin2023origin} generates trajectories in the format of grids, by a condition model guided by the origin-destination information.

\vspace{-0.2cm}
\subsection{Imputation}\label{subsec:imputation}

In the domain of time series and spatio-temporal data analysis, imputation refers to generating the unobserved data conditioned on the given observed data. CSDI \cite{tashiro2021csdi} proposes a score-based diffusion model that probabilistically imputes missing time series and spatio-temporal data. MIDM \cite{wang2023observed} redefines the evidence lower bound (ELBO) for conditional diffusion models tailored for multivariate time series imputation, which ensures the consistency of observed and missing values. PriSTI \cite{liu2023pristi} introduces a conditional diffusion framework specifically designed for spatio-temporal data imputation, using global context priors and geographic relationships to tackle scenarios with high missing data rates due to sensor failures. More recently, TabCSDI \cite{zheng2022diffusion} and MissDiff \cite{ouyang2023missdiff} leverage diffusion models for imputing missing values in tabular data, each addressing distinct aspects of this complex challenge. TabCSDI manages categorical and numerical data effectively, providing a tailored solution to diverse data types common in tabular datasets. On the other hand, MissDiff enhances the diffusion model's training process by introducing a novel masking technique for regression loss, which ensures consistent learning of data distributions and robustness against different missing data scenarios.

\vspace{-0.2cm}
\subsection{Anomaly Detection}\label{subsec:anomaly-detection}

Anomaly detection aims to identify anomalies in the given time series or spatio-temporal data, which is quite practical and crucial for many real-world applications. A majority of works have been proposed for general time series anomaly detection. Initially, DiffAD \cite{xiao2023imputation} and ImDiffusion \cite{chen2023imdiffusion} both explored the synergy of imputation techniques with diffusion models for time series anomaly detection, enhancing the robustness of anomaly detection processes by accurately modeling complex dependencies.  Concurrently, \cite{pintilie2023time} and \cite{livernoche2023diffusion} use similar diffusion-based methodologies to address anomaly detection but apply different enhancements to improve performance and computational efficiency. Meanwhile, \cite{wang2024drift} and DDMT \cite{yang2023ddmt} tackled the specific challenges of instability and noise, implementing advanced diffusion reconstruction techniques to maintain accuracy in dynamic environments.

Besides the above methods designed for general time series anomaly detection. There are other works that apply advanced diffusion techniques across various domains. For instance, Maat \cite{lee2023maat} anticipates performance metric anomalies in cloud services using a conditional denoising diffusion model to forecast metrics and detect anomalies. \cite{mueller2024attention} enhances data synthesis for machine fault diagnosis using an attention mechanism within a conditional diffusion model framework. Diffusion-UDA \cite{zhao2024diffusion} proposes a diffusion-based method for unsupervised domain adaptation in submersible fault diagnosis, leveraging diffusion processes to adapt domains for effective fault recognition.

\vspace{-0.2cm}
\section{Data Perspective} \label{sec: Modality} 
In this section, we analyze current work from a data perspective, including time series and spatio-temporal data, as shown in~\shortautoref{fig:data illustration}. We focus on introducing how diffusion models capture the unique properties of different data modalities.

\vspace{-0.2cm}
\subsection{Time Series}\label{subsec:time-series-data}

Time series analysis is a critical issue explored across various real-world scenarios, including retail sales forecasting, filling in missing data in economic time series, identifying anomalies in industrial maintenance, and categorizing time series from different domains.

\subsubsection{Univariate Time Series}
Univariate time series is characterized by having only one variable of interest observed over a period of time. Data in this category includes ECG signal, audio,  electricity load, etc. The complex sequential patterns (e.g., the trend and the periodicity) are the core data property in univariate time series, which pose great challenges in developing relevant diffusion models.

Diffusion models for univariate time series are primarily developed to model the uncertainty present in the data, facilitating tasks such as probabilistic forecasting or data generation. An essential element within the diffusion model is the denoising network, which determines the extent of noise removal at each stage. In the context of univariate time series, the denoising network commonly employs the 1-D CNN to identify sequential patterns in the input data. For instance, both WaveGrad~\cite{chen2020wavegrad} and DiffWave~\cite{kong2020diffwave} incorporate a 1-D CNN into the denoising network to extract local sequential features from audio sequences. Furthermore, some studies opt to transform the univariate time series into the frequency domain initially to enhance the capture of long-term sequential correlations from a global standpoint. This transformation converts the time series from 1-D to 2-D data, enabling the use of a 2-D CNN in the denoising network to capture correlations within the frequency domain, such as \cite{koa2023diffusion} for the stock price forecasting and DiffLoad \cite{wang2023diffload} for electricity load forecasting.

\subsubsection{Multivariate Time Series}
Multivariate time series is characterized by having multiple variables of interest observed over the same period of time. Instead of the sequential patterns in each variate, multivariate time series also leverage the interdependencies among different variables to capture more comprehensive information for downstream tasks. 

Beyond approaches that focus on univariate time series, there have been efforts towards multivariate time series. Multivariate time series is naturally a 2-D data similar to image data, where the signal of each variate can be considered as one channel in the image. To this end, diffusion models for multivariate time series usually adopt the vanilla U-Net as the denoising net, which is a common practice in diffusion models for images. 

For example, \cite{durall2023deep} first transforms demultiplexing, denoising, and interpolation into image-to-image transformation tasks, then leverages the diffusion model for different tasks. In the demultiplex task, the diffusion model learns to remove the multiples without removing the primary energy. In the denoising and interpolation task, the diffusion model learns to eliminate undesired uncorrelated noise and recover the missing values, while preserving the inherent characteristics of the data. MIDM \cite{wang2023observed} proposes a novel multivariate imputation diffusion model that incorporates correlations between observed and missing values. By re-deriving the ELBO of the conditional diffusion model, MIDM ensures consistency between observed and missing data, thus leading to improved imputation accuracy. Diff-E \cite{kim2023brain} further utilizes the diffusion model for representation learning, which combines DDPMs with a conditional autoencoder to enhance the decoding performance of speech-related EEG signals. 

\vspace{-0.2cm}
\subsection{Spatio-Temporal Data}\label{subsec:spatio-temporal-data} 

In real-world systems,  a myriad of elements interact with each other both spatially and temporally, resulting in a spatio-temporal composition. Spatio-temporal data (STD) is the de facto most popular data structure for injecting such structural information into the formulation of practical problems. In this section, we introduce developments of diffusion models for spatio-temporal data, mainly with two data modalities: (1) the Spatio-Temporal Graph (STG) and (2) the Spatio-Temporal Trajectory (STT). Unlike diffusion models for time series data, existing works in STD need to model the temporal dependencies as well as the spatial dependencies, making the task even more challenging. 

\subsubsection{Spatio-Temporal Graph}
The spatio-temporal graph is generated by sensors at different places in a period of time, where the correlation of those sensors is typically described as a graph. We categorize diffusion models for spatio-temporal graphs into domain-oriented and domain-agnostic works. 

Most of the existing works fall into the category of domain-oriented spatio-temporal graph diffusion models. They typically leverage the powerful distribution learning abilities of diffusion models for spatio-temporal data mining tasks in specific domains, such as traffic and climate.  In this research trend, the majority of works come from the traffic domain. DiffUFlow \cite{zheng2023diffuflow} represents pioneering efforts as traffic diffusion models, which convert fine-grained urban flow inference into a denoising diffusion process. SpecSTG \cite{lin2024specstg} advances the field by conducting the diffusion process in the spectral space projected by the STG, resulting in faster inference speeds. In a similar vein, \cite{liang2023dynamic} uses a diffusion model to infer the link probability and reconstruct causal graphs in the decoder stage adaptively for the STG forecasting task. Furthermore, \cite{lin2023origin} solves the origin-destination travel time estimation problem with a two-step process, where the first step predicts the possible travel route with a diffusion model conditioned on the given origin-destination pair. Beyond the traffic domain, there have been initiatives towards the climate. A notable example is \cite{chen2024quantifying}, which quantifies the uncertainty of air quality forecasting based on the spatio-temporal diffusion model. Similarly, SRNDiff \cite{ling2024srndiff} conducts the short-term rainfall nowcasting through the conditional diffusion model.

In the second group, researchers focus on developing domain-agnostic models, which can achieve promising performance across a variety of domains. For instance, DiffSTG represents a pioneering effort by introducing a unified diffusion framework for multiple STG tasks, such as forecasting and imputation. Concurrently, PriSTI \cite{liu2023pristi} and \cite{yun2023imputation} study the spatio-temporal imputation task while treating the task of recovering unobserved values as a denoising process conditioned on the observed values. Similarly, DYffusion \cite{cachay2023dyffusion} models the temporal dynamics directly within diffusion steps, leading to a stochastic, time-conditioned forecasting network. Another noteworthy contribution is \cite{hu2023towards} under the domain-agnostic and task-agnostic setting, which aims to harness the power of diffusion models, and unify diffusion models for probabilistic spatio-temporal graph learning.

\subsubsection{Spatio-Temporal Trajectory}

Spatio-temporal trajectory is a sequence of locations ordered by time that describes the movements of an object in a geographical space. Analysis of spatio-temporal modality is particularly crucial to discovering the mobility patterns of moving objects, which serves as the foundation for many downstream tasks, such as POI recommendation and next-location forecasting. In terms of diffusion models for trajectory data, most works are developed to solve the trajectory generation task. They leverage the ability of the diffusion model to learn high-dimensional data distribution while injecting spatio-temporal correlation into the diffusion process. As exemplified by DiffTraj \cite{zhu2024difftraj}, which generates the GPS trajectory with the diffusion probabilistic model in an unconditioned manner. Similarly, \cite{chu2024simulating} utilizes the diffusion model for generating high-quality human mobility data. More recently, Diff-RNTraj \cite{wei2024diff} further generates the trajectory constrained by the road-network. Beyond generation, there is a growing interest in applying diffusion models for trajectory forecasting tasks, and \cite{Yao2023CSIS} and \cite{liu2024intention} exemplify this trend.

\vspace{-0.3cm}
\section{Application Perspective} \label{sec:applications} 

\begin{figure}
    \centering
    \includegraphics[width=0.85\linewidth]{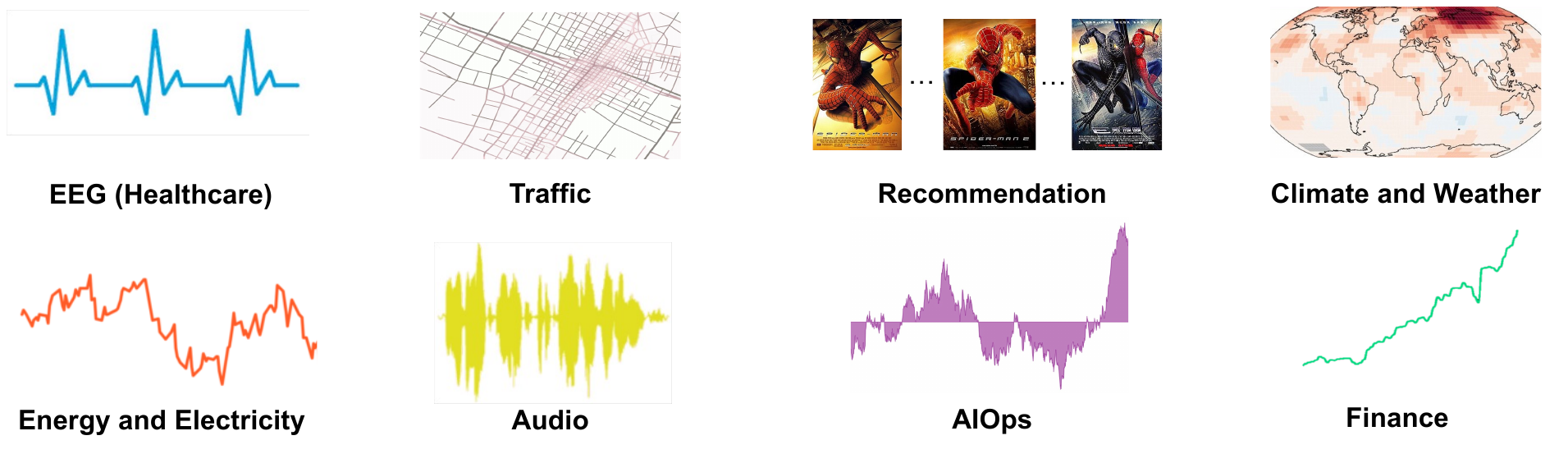}
    \vspace{-0.2cm}
    \caption{Examples of time series and spatio-temporal data for different application scenarios.} \vspace{-0.4cm}
    \label{fig:application data}
\end{figure}

In this section, we summarize and discuss the most important applications of diffusion models in time series and spatio-temporal data, including healthcare, smart city, recommendation, climate and weather, energy and electricity, audio, video, and so on. The toy signal examples of each application are shown in ~\shortautoref{fig:application data}.

In detail, we will discuss the models in different applications, aiming to show light on the model design of practical applications.

\vspace{-0.3cm}
\subsection{Healthcare}
In recent years, diffusion models have emerged as a powerful class of generative models with significant implications for time series in healthcare. These models, known for their ability to generate high-fidelity samples through a process of gradual refinement, have found diverse applications ranging from the synthesis of electronic health records (EHRs) to the enhancement of biomedical signal analysis. The following are some main topics in this area. EEG Signal Synthesis and Enhancement: \cite{neifar2023diffecg} introduces a generalized probabilistic diffusion model tailored for the synthesis of electrocardiogram (ECG) signals, aiming to support cardiac health research and diagnostic training with realistic data generation. \cite{aristimunha2023synthetic} demonstrates the potential of diffusion models to create synthetic EEG signals, facilitating sleep disorder studies and neurological research. \cite{li2023descod,alcaraz2023diffusion} focus on enhancing the quality of ECG signals, addressing issues like baseline wander and noise, and generating conditional ECG signals based on specific patient states. Healthcare Data Augmentation and Synthesis: \cite{shu2023data} highlights the use of diffusion models to augment datasets for seizure forecasting, thereby improving the robustness and accuracy of predictive models. \cite{yuan2023ehrdiff,ceritli2023synthesizing} explore the synthesis of realistic EHRs, tackling the challenges of data privacy and scarcity by providing synthetic datasets for research and model training. Diagnostic and Predictive Analytics: \cite{li2023generative} leverages generative models to enhance the detection of autism, showcasing the broad applicability of these models in diagnosing and understanding complex conditions. In~\cite{chang2024transformer}, diffusion models are applied to forecast critical physiological parameters in ICUs, demonstrating their potential to save lives by predicting adverse events before they occur. Novel Methodologies and Techniques: \cite{zhao2024multi} exemplifies the versatility of diffusion models in handling multiple tasks simultaneously, such as predicting mortality in critically ill patients, showcasing the potential of comprehensive care management systems.
 
\vspace{-0.3cm}
\subsection{Traffic} 
There are also many works for traffic applications using diffusion models. The following are some of them. DiffUFlow~\cite{zheng2023diffuflow} employs a denoising diffusion model to enable the model to capture the stochastic nature of urban flows and generate accurate and fine-grained forecasting. DiffTraj~\cite{zhu2024difftraj} presents a groundbreaking approach to generating GPS trajectories using a spatio-temporal diffusion probabilistic model. The model works by reconstructing geographic trajectories from white noise through a reverse trajectory denoising process, effectively turning random noise into meaningful trajectory data that reflects real-world movement patterns. DVGNN architecture~\cite{liang2023dynamic} comprises an encoder component that learns latent node embeddings through GCN layers and a decoder component that treats the relationships of latent states as a diffusion process subject to stochastic differential equations. This allows for the inference of internal causal relationships among neighbor nodes and the formation of dynamic causal graphs. \cite{guimaraes2023predicting} addresses the growing risk of collisions between space objects by developing a diffusion model that predicts the positional uncertainty of objects during close encounters. \cite{li2024difftad} introduces a model that utilizes denoising diffusion probabilistic models for identifying anomalies in vehicle trajectories, aimed at improving the detection of unusual patterns in vehicle movements. SpecSTG~\cite{lin2024specstg} aims to efficiently handle the complexities of spatio-temporal data, offering a probabilistic approach to forecasting traffic flows. This method is distinguished by its speed and accuracy in generating forecasts and is a promising solution for real-time traffic management and planning applications. 

\vspace{-0.3cm}
\subsection{Sequential Recommendation}

In the scenario of sequential recommendation, diffusion models are also widely applied. \cite{liu2023diffusion} and \cite{wang2023diffusion} both explore the application of diffusion models in enhancing sequential recommendation systems, yet they focus on different aspects and methodologies within this domain. That is, \cite{liu2023diffusion} aims to bridge the gap between discrete item identities and the continuous nature of the data generated by the diffusion models, while \cite{wang2023diffusion} focuses on predicting users’ future interaction probabilities by corrupting and then denoising their interaction histories and addressing challenges specific, such as high resource costs and temporal shifts in user preferences. DCDR~\cite{lin2023discrete} is a discrete forward process with tractable posteriors and a conditional reverse process tailored for sequence generation. RecFusion~\cite{benedict2023recfusion} introduces a binomial diffusion process tailored for one-dimensional data, such as time series of user interactions, showcasing its effectiveness in sequential recommendation scenarios. DiffuRec~\cite{li2023diffurec} proposes a specific diffusion model framework designed for sequential recommendation, focusing on efficiently capturing and predicting evolving user interests. Diffusion models have shown significant promise in improving sequential recommendation systems. Their ability to simulate the spread of information and preferences offers a nuanced method for predicting user behavior. Future research could explore hybrid models, scalability, and real-time adaptation, further enhancing the relevance and personalization of recommendations.

\vspace{-0.3cm}
\subsection{Climate and Weather}
Diffusion models, initially developed for image generation, have found novel applications in weather forecasting due to their ability to generate high-resolution, realistic outputs. These models work by gradually refining a signal from a random noise distribution towards a desired output, making them well-suited for predicting complex atmospheric phenomena. \cite{leinonen2023latent} focuses on precipitation nowcasting with an emphasis on accurate uncertainty quantification, leveraging latent space representations to efficiently model and predict rainfall intensity and distribution. \cite{kurinchiwirediff} presents a wind resolution-enhancing model that improves the detail and accuracy of wind speed forecasts by refining coarse forecasting to higher resolutions. SwinRDM~\cite{chen2023swinrdm} integrates Swin Recurrent Neural Networks (SwinRNN) with diffusion models for state-of-the-art weather forecasting, achieving high resolution and quality by capturing both spatial and temporal dynamics. DiffESM~\cite{bassetti2023diffesm} utilizes diffusion models for the conditional emulation of Earth System Models (ESMs), offering a computationally efficient alternative to traditional simulation techniques with improved accuracy. \cite{hatanaka2023diffusion} addresses the challenge of high-resolution solar forecasting, showcasing the model's ability to accurately predict solar irradiance variations. DiffAD~\cite{huang2024diffda} introduces a diffusion model-based approach for weather-scale data assimilation, enhancing the integration of observational data into forecast models for improved accuracy.

\vspace{-0.3cm}
\subsection{Energy and Electricity}
Diffusion models are now being adapted to tackle challenges in energy systems, offering novel solutions for probabilistic forecasting, signal synthesis, and system security. DiffLoad~\cite{wang2023diffload} is a novel approach to load forecasting that leverages diffusion models to quantify uncertainties effectively. By incorporating uncertainty quantification, DiffLoad offers a more reliable forecast, which is crucial for grid stability and operational planning. \cite{wang2023customized} explores how conditional diffusion models can synthesize load profiles for individual electricity customers, enhancing the personalization of demand-side management strategies. DiffPLF~\cite{li2024diffplf} addresses the challenges of load forecasting by providing probabilistic forecasts of EV charging loads, facilitating better grid management and infrastructure planning. \cite{capel2023denoising} discusses the application of denoising diffusion probabilistic models in energy forecasting, emphasizing the model's ability to handle uncertainty and provide probabilistic forecasts, which are essential for integrating renewable energy sources. \cite{dong2023short} demonstrates how conditional latent diffusion models can be used to generate realistic short-term wind power scenarios, aiding in system operation and planning.

\vspace{-0.3cm}
\subsection{Audio}
Diffusion models have also found promising applications in generating waveforms, synthesizing audio, music generation and enhancing speech, offering significant improvements in quality, realism, and control over the generated or enhanced audio. Waveform Generation and Audio Synthesis: WaveGrad~\cite{chen2020wavegrad} introduces a gradient-based approach to generate high-fidelity waveforms, demonstrating the potential of diffusion models in audio synthesis without the need for autoregressive models. DiffWave~\cite{kong2020diffwave} extends the capabilities of diffusion models in audio synthesis, showcasing their versatility across different audio synthesis tasks, including voice and music. Music Generation: \cite{zhang2023fast} explores the fusion of diffusion models and GANs for generating symbolic music, allowing for emotion-driven control over the generation process. DiffuseRoll~\cite{wang2023diffuseroll} highlights the application of diffusion models in multi-track music generation, providing nuanced control over various musical attributes. Speech Enhancement: \cite{richter2023speech} introduces a diffusion-based generative approach to simultaneously enhance speech and reduce reverberation, showcasing significant improvements over traditional methods. \cite{serra2022universal} focuses on the universal applicability of diffusion models and demonstrates effectiveness in enhancing speech across a wide range of conditions. \cite{lay2023reducing} proposes a method to reduce the prior mismatch in stochastic differential equations, enhancing model performance in speech tasks. CRA-DIFFUSE~\cite{qiu2023cra} introduces a pre-denoising step in the time-frequency domain to improve cross-domain speech enhancement, illustrating the potential for methodological innovations within diffusion model applications. The application of diffusion models in audio processing has opened new avenues for research and development, offering novel solutions to longstanding challenges. As the field progresses, future research may focus on improving model efficiency, reducing computational demands, and exploring untapped applications within audio processing.

\vspace{-0.3cm}
\subsection{AIOps}
Cloud computing's reliability is paramount, yet it faces challenges like performance anomalies, unpredictable network traffic, and incomplete data. Diffusion models have emerged as a powerful tool to tackle these issues, offering new approaches for predictive maintenance, realistic traffic simulation, and enhanced failure forecasting. Maat~\cite{lee2023maat} applies conditional diffusion models to anticipate performance metric anomalies in cloud services, demonstrating the potential for early anomaly detection, potentially reducing downtime, and improving service reliability. NetDiffus~\cite{sivaroopan2023netdiffus} applies diffusion models to time-series imaging for generating realistic network traffic patterns, showing how simulated traffic can support network planning, testing, and anomaly detection, enhancing network management and security. \cite{yang2023diffusionaiops} investigates diffusion models for imputing missing data in time series, aiming to improve cloud failure forecasting accuracy, highlighting the effectiveness of diffusion models in handling data gaps, leading to better predictive outcomes for cloud service failures.

\vspace{-0.3cm}
\subsection{Finance}
The complexity and stochastic nature of financial markets make them an ideal candidate for the application of diffusion models, which can capture non-linearities and intricate patterns in data. Recent advancements have highlighted the potential of diffusion models in enhancing stock price forecasting, generating realistic financial tabular data, and augmenting stock factor data for improved investment strategies. \cite{koa2023diffusion} presents a novel approach to predict stock prices using a Diffusion Variational Autoencoder, aiming to better capture the stochastic nature of the market, highlighting the model's ability to handle market volatility. FinDiff~\cite{sattarov2023findiff} focuses on generating synthetic financial tabular data to address the scarcity of publicly available financial datasets. The synthetic data generated by FinDiff closely mimics real financial datasets, potentially aiding in model training and regulatory compliance without compromising sensitive information. DiffSTOCK~\cite{daiya2024diffstock} employs diffusion models for probabilistic relational reasoning in stock market forecasting, aiming to capture the complex interdependencies between different market factors.

\vspace{-0.3cm}
\section{Outlook and Future Opportunities} \label{sec:future directions} 

In this section, we point out some future research directions of diffusion models for time series and spatio-temporal data that are worthy of further investigation. (1) \textbf{\textit{Scalability and Efficiency}}: Diffusion models remain computationally demanding, limiting their use in resource-constrained settings. Future work should focus on developing lighter architectures, enhancing sampling, model compression, and parallelization to improve real-world applicability. (2) \textbf{\textit{Robustness and Adaptability}}: Real-world time series often contain noise, missing values, anomalies, and distribution shifts. Strengthening the robustness and generalization of diffusion models across datasets and scenarios, as well as enabling dynamic adaptation to changing environments, is essential. (3) \textbf{\textit{Domain-Generalizable Challenges}}: Time series benchmarks are typically small and exhibit strong autocorrelation, causing overfitting. Despite the regularization effect of stochastic denoising, generalization to unseen domains remains limited. The recently identified "accuracy law"~\cite{wang2025accuracy} suggests that many benchmarks may approach their intrinsic predictability limits, highlighting risks of saturation. Future work should investigate domain-generalizable diffusion frameworks via temporal augmentation, cross-domain transfer, and prior-guided regularization, advancing toward foundation models for heterogeneous temporal modalities. (4) \textbf{\textit{Prior Knowledge Guided Generation}}: Generated time series, trajectories, and spatio-temporal patterns should satisfy domain-specific constraints (e.g., road networks, population dynamics, thermodynamics), yet current diffusion models insufficiently incorporate such prior knowledge. (5) \textbf{\textit{Multimodal Data Fusion}}: Time series often co-occur with text or images. Integrating multimodal information can enhance analysis in applications such as finance and healthcare. Thus, new architectures are needed for effective multimodal fusion. (6) \textbf{\textit{Emerging Trends and Cross-Domain Inspirations}}: Advances in image-based diffusion, such as latent diffusion~\cite{rombach2021highresolution}, consistency models, and rectified flows~\cite{song2023consistency,armegioiu2025rectified}, offer efficiency and controllability insights, though direct transfer is hindered by temporal causality and non-stationarity. Adapting these ideas to time-aware diffusion and exploring multimodal conditioned temporal generation represent key directions. (7) \textbf{\textit{Integration of LLMs and Diffusion Models}}: Combining LLMs with diffusion models may enhance temporal reasoning and support more comprehensive decision-making by uniting semantic understanding with generative modeling. Future work could develop hybrid architectures leveraging both capabilities.

\vspace{-0.3cm}
\section{Conclusion} \label{sec:conclusion}

In this survey, we presented a comprehensive overview of the advancements and applications of diffusion models in the context of time series and spatio-temporal data analysis. We categorized diffusion models into unconditioned and conditioned types, each offering distinct advantages and challenges. Furthermore, we examined the various tasks associated with these models, including forecasting, generation, imputation, and anomaly detection. Additionally, we explored different application scenarios and provided insights into future opportunities and directions in this research field. It is our hope that this survey will contribute to the advancement of research in the area of diffusion models for time series and spatio-temporal data analysis.

\bibliographystyle{ACM-Reference-Format}
\bibliography{sample-base}

\end{document}